\documentclass[11pt]{article}

\usepackage{acl}
\usepackage{times}
\usepackage{latexsym}
\usepackage[T1]{fontenc}
\usepackage[utf8]{inputenc}
\usepackage{microtype}
\usepackage{inconsolata}
\usepackage{graphicx}
\usepackage{amsmath,amssymb}
\usepackage{adjustbox}
\usepackage{booktabs} 
\usepackage{multirow} 
\usepackage{arydshln}
\usepackage{xcolor}
\usepackage{colortbl}
\usepackage{tikz}
\usepackage{cleveref}
\usepackage{float}
\usepackage[most]{tcolorbox}

\definecolor{green1}{HTML}{02C9AF}
\definecolor{LightGreen}{HTML}{DFF6DD}
\definecolor{LightYellow}{HTML}{FFF3CD} % Bootstrap 风格浅黄（可读性好）
\definecolor{LightRed}{HTML}{F8D7DA}    % Bootstrap 风格浅红

\definecolor{DeltaPos}{RGB}{0,128,0}      % green
\definecolor{DeltaNeg}{RGB}{200,0,0}      % red
\definecolor{DeltaNeu}{gray}{0.45}        % gray

\definecolor{lightblue}{rgb}{0.8,0.85,1}
\definecolor{lightred}{rgb}{1, 0.8, 0.8}

\title{Beyond Uniform Tokens: Adaptive Compression for Time Series \\ Language Models}

\author{
 \textbf{Jialin Gan}\textsuperscript{1}\thanks{Equal contribution.}
 \quad
 \textbf{Xin Qiu}\textsuperscript{1}\footnotemark[1]
 \quad
 \textbf{Guangzhe Chen}\textsuperscript{2}
 \quad
 \textbf{Xue Wang}\textsuperscript{3}
\\
 \textsuperscript{1}Zhejiang University
 \quad
 \textsuperscript{2}Harbin Institute of Technology
 \quad
 \textsuperscript{3}Shandong University
}

\begin{document}
\maketitle
\begin{abstract}
% Large Language Models (LLMs) have recently shown strong potential for time series (TS) analysis by integrating numerical signals with rich text prompts. 
% In this process, although TS are encoded into token sequences and fed into LLMs in a manner similar to text prompts, TS tokens differ from prompt tokens in their underlying structure and semantics, a distinction that remains underexplored in existing LLM-based TS systems. 
% In this work, we investigate both TS tokens and prompt tokens from a new perspective. 
% First, we re-examine TS tokens in the frequency-domain and reveal that token-level information contribution is highly uneven: large groups of tokens share redundant spectral patterns, while a small subset acts as critical frequency anchors. 
% Motivated by this observation, we introduce a frequency-based token merging strategy for TS tokens, which preserves essential information while eliminating redundancy.
% Second, we uncover a unexplored pyramidal decay effect in prompt tokens, showing that their influence rapidly diminishes with model depth.
% Building on these findings, we propose \textbf{\textit{TokenDecouple}}, a unified framework that decouples the compression of TS tokens and prompt tokens. Extensive experiments across diverse TS tasks show that TokenDecouple achieves up to \textit{\textbf{7.68×}} inference speedup while delivering performance improvements in \textit{\textbf{78\%}} of evaluated scenarios, establishing an efficient paradigm for LLM-based TS analysis.
Large language models (LLMs) have enabled time series (TS) analysis by jointly modeling numerical observations and textual context through a shared token interface. However, TS tokens and prompt tokens exhibit fundamentally different information structures, making uniform token processing inefficient. In this paper, we study token efficiency in TS language modeling from an asymmetric-token perspective. We show that TS tokens have highly uneven spectral contributions, where many tokens share redundant frequency patterns while a small subset preserves critical temporal evidence. We also observe that prompt-token influence attenuates with model depth, suggesting that full prompt retention across all layers is unnecessary. Based on these findings, we develop an adaptive token budgeting framework that compresses TS tokens via frequency-domain structure and progressively reduces prompt tokens across layers. Experiments across forecasting, classification, imputation, and anomaly detection demonstrate up to \textit{\textbf{7.68$\times$}} inference acceleration and performance gains in \textit{\textbf{78\%}} of evaluated settings, showing the effectiveness of asymmetric token compression for scalable TS foundation models.

\end{abstract}
\section{Introduction}
Large Language Models (LLMs) have reshaped the landscape of NLP, exhibiting remarkable proficiency in understanding and reasoning~\cite{roy2024exploring,qin2025largelanguagemodelsmeet,matarazzo2025surveylargelanguagemodels}. As models scale in both capacity and data diversity, their influence has extended well beyond language, enabling powerful multimodal modeling that unify information from audio, images, and video~\cite{han2024multimodal,zhang2025llava,guo2025aligned}. Therefore, a growing body of research integrates LLMs into cross-modal modeling between text and time series (TS), enabling strong performance on event forecasting, causal reasoning, anomaly detection~\cite{wang2024news,pan2024s,chang2025llm4ts}. For example, CrisisTS aligns crisis-related textual reports with meteorological TS signals to improve multilingual urgency assessment~\cite{meunier2025crisists}. Time-MQA supports multi-task TS question answering (QA) by coupling language reasoning with continual pre-training on temporal data~\cite{kong2025time}. ITFormer provides an end-to-end pipeline that connects TS modeling with language generation, enabling  multimodal QA for engine-maintenance scenarios~\cite{wang2025itformer}.

Although LLMs have begun to show encouraging progress on TS tasks, the nature of temporal signals remains different from language and is still underexplored. 
TS are highly structured, exhibit strong temporal continuity, and often undergo pronounced distribution shifts in real-world environments~\cite{he2024robust,lu2025nonstationarytimeseriesforecasting}. 
Different frequency components of TS become dispersed in the time domain, yet LLMs lack mechanisms for recognizing these distinct frequency roles~\cite{zhou2022fedformer}. This mismatch obscures the true functional contribution of each token and complicates efficient temporal representation learning. {{Therefore, a key open problem is how to effectively uncover the deeper information embedded within TS tokens, which is both challenging and essential.}}

To address this issue, it is essential to re-examine TS tokens from a frequency-domain perspective.
Our analysis reveals several key mechanisms. Specifically, large groups of tokens share highly similar spectral patterns within certain frequency bands, which manifest inside the model as nearly indistinguishable attention behaviors. In contrast, only a small subset of tokens carries dominant information at specific frequencies. Motivated by these observations, \emph{we introduce the first token-level frequency-based merging strategy for LLM-based TS systems}. By isolating true information-carrying “executors” and removing redundancy.

In addition, TS are highly abstracted: a local blackout may manifest only as a sudden drop or near-zero collapse in electricity consumption, and a scandal involving a CEO as a downward trend in stock price. Because such rich causes collapse into simple numerical patterns, LLM-based TS systems typically rely on substantial text prompts to provide contextual guidance. These prompts both compensate for the compressed nature of TS data and help activate the model’s reasoning ability~\cite{jin2023time,gruver2023large}. However, introducing large numbers of prompt tokens inevitably incurs substantial computational overhead~\cite{dao2022flashattention,kwon2023efficient}. This trade-off gives rise to a central challenge: how to reduce token usage but preserving essential information.

Building on this insight, \emph{We reveal a surprisingly \textbf{“pyramidal decay”} effect in prompt tokens}: their influence diminishes rapidly as model depth increases. Motivated by this observation, we introduce a prompt token compression strategy that aggressively reduces prompt tokens in early layers. Moreover, we find that textual information is fully absorbed and transferred to TS tokens within only a few shallow layers (at most three layers), indicating that prompt tokens need not propagate through the entire model. As a result, the substantial computational cost can be drastically reduced with negligible impact on model performance.
% 这里可以留一段 写效果
Our contributions can be summarized as follows:
\begin{enumerate}
    \item We provide \textit{the first systematic frequency-domain perspective on TS tokens}, revealing how different frequency components contribute unevenly to token-level information.

    \item We uncover \textit{a pyramidal decay effect in prompt tokens}, showing that their influence rapidly diminishes with model depth and can be safely compressed in early layers.

    \item We propose \textit{\textbf{TokenDecouple}}, \textit{a unified framework for decoupled compression of TS and prompt tokens}, enabling substantial efficiency gains while even improving performance.
\end{enumerate}

\section{Related Work}
% \vspace{-0.3cm}
\textbf{LLMs for TS Analysis.}\quad
TS analysis covers a wide range of tasks, including forecasting, anomaly detection, classification, and question-answering (QA)~\cite{xie2024chatts,liu2025timecma,liu2025calf}. Recently, LLMs have been explored as a general-purpose solution for them. A direct way is to feed raw TS points into the LLM, but it fails to bridge the modality gap between TS and text~\cite{gruver2023large,liu2024lstprompt}. Another line of work converts TS into images and relies on vision language Models (VLMs), but they often struggle with fine-grained temporal patterns~\cite{zhong2025time,ruan2025vision}. Consequently, the dominant approach is to embed TS into representation space, and then align TS tokens with text tokens to construct LLM-based TS analysis systems. For example, Time-MMD~\cite{liu2024time}, GPT4MTS~\cite{jia2024gpt4mts}, ChatTS~\cite{xie2024chatts}, and TimerBed~\cite{kong2025time} combine TS data with domain knowledge to support TS tasks. Because TS data are already irreversibly compressed during collection, extensive prompts are required to activate LLMs’ prior knowledge, but combining them with high-dimensional TS tokens incurs prohibitive computational costs.

\smallskip
\noindent\textbf{TS Representation Analysis.}\quad
Compared with the relatively sparse spatial distribution of text, TS exhibits a more continuous structure and more concealed semantic features~\cite{zhang2025timeseriesanalysisfrequency,yue2025freeformerfrequencyenhancedtransformer}. In theory, the types of TS patterns are nearly impossible to enumerate, with semantic information hidden within complex trends, cycles, and fluctuations~\cite{li2025ftmixer,qin2025sfdformer}. While some studies have explored TS from various perspectives, deep analyses remain rare~\cite{yuan2024d,he2025unified}. One key reason is that most existing analyses focus on the time-domain, where multiple semantic factors are entangled within continuous signals, making systematic characterization difficult~\cite{wu2022timesnet}. In contrast, the frequency-domain distribution of TS is more concentrated and discretized. Motivated by these challenges, we conduct a frequency-domain analysis of TS token in LLM-based TS analysis systems, revealing their distribution characteristics and modes of information contribution.

\section{Preliminary}
\begin{figure*}[t]
  \includegraphics[width=\textwidth]{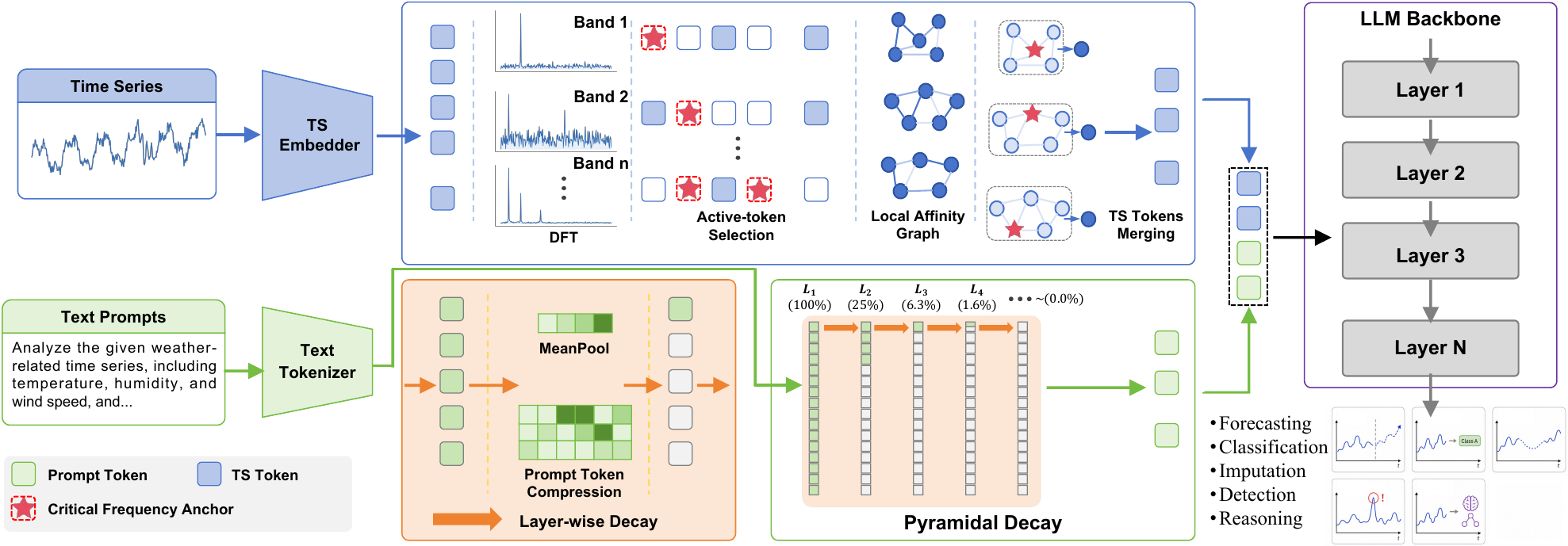}
  \caption{
Overview of \textbf{\textit{TokenDecouple}}. 
We decouple token compression for TS and prompt tokens by merging redundant TS tokens in the frequency-domain and reducing prompt tokens under the pyramidal decay effect.
}
\label{fig:main}
\end{figure*}
\subsection{LLM-based TS Analysis System}
A standard LLM-based TS analysis system consists of four components: a TS embedder, a text tokenizer, an LLM backbone, and a task-specific head. The TS embedder transforms raw TS signals into TS tokens, and the text tokenizer maps text prompts into prompt tokens. These two token streams are then fused within the LLM backbone, which produces the final outputs through a task-specific head. However, because the LLM backbone is pre-trained exclusively on text data, a fundamental distribution shift exists between TS and language. As a result, an effective LLM-based TS analysis system must establish a robust alignment between TS data and text data. 
Existing approaches follow two major directions: \textit{\textbf{(1) input-space alignment}}, where TS–text reprogramming is performed before tokens enter the LLM. This reshapes TS tokens into a language-compatible format, allowing them to interact effectively with prompt tokens~\cite{jin2023time}; and \textit{\textbf{(2) latent-space alignment}}, where the distribution mismatch is addressed by lightly fine-tuning a small subset of LLM parameters~\cite{zhou2023one,tang2025llm,chang2025llm4ts}. 
It reduces modality drift while minimizing catastrophic forgetting~\cite{french1999catastrophic}.

\subsection{TS Tasks: From Patterns to Reasoning}
\label{TS Tasks: From Patterns to Reasoning}
In real-world, TS reflect phenomena with varying levels of complexity.
This natural hierarchy motivates separating TS analysis into two major groups. The first consists of \emph{\textbf{pattern-centric tasks}}, where solving the problem depends primarily on understanding the observable structure of the sequence. The second consists of \emph{\textbf{reasoning-centric tasks}}, where the model must infer unobserved generative factors beyond the visible temporal patterns. 

\textbf{Pattern-centric tasks} aim to interpret observable TS patterns and follow a natural progression of difficulty from (i)imputation to (ii)classification, (iii)anomaly detection, and (iv)forecasting. Interpolation primarily relies on strong local continuity and requires minimal reasoning.
Classification depends on holistic and multi-scale representations to capture class-specific patterns.
Anomaly detection requires modeling normality to identify deviations.
Forecasting is the most demanding, as it requires generating unobserved future values while capturing trends, periodicity, and short-term dynamics.

\textbf{Reasoning-centric tasks} move beyond pattern recognition and require the model to infer the causal structure, categorical logic, latent events, or unobserved factors. (i) Simple deterministic reasoning represents the easiest level, where decisions are governed by single-cue determination and rely on salient local patterns. (ii) Complex deterministic reasoning is harder, as outcomes depend on the joint configuration of multiple physiological or domain-specific features rather than any single cue. (iii) Probabilistic reasoning is the most challenging, requiring the model to handle probability distributions, ambiguity, and latent variables.

\begin{figure}[t]
  \includegraphics[width=\linewidth]{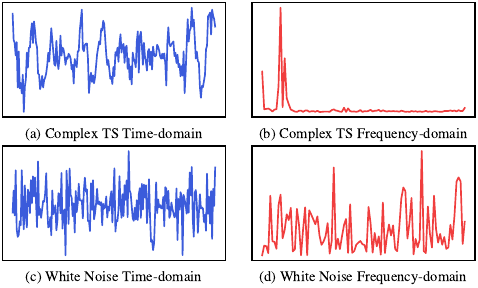}
  \caption{The complex TS appears entangled in the time domain but shows concentrated energy in a few frequency bands, while white noise has a nearly uniform spectrum, indicating poor compressibility.}
\label{fig:time_vs_frequency_comparison}
\end{figure}
\section{TokenDecouple: Decoupled Compression of TS and Prompt Tokens}
As shown in Fig.~\ref{fig:main}, \textbf{\textit{TokenDecouple}} decouples token compression for TS tokens and prompt tokens. 
Specifically, it performs frequency-domain merging for TS tokens, while applying pyramidal compression to prompt tokens to reduce redundant textual overhead in shallow layers.

\subsection{Two Approaches: Time-Domain vs. Frequency-Domain} 
Due to the large computational burden imposed by TS tokens, effective compression is essential for efficient TS analysis. 
Token merging has emerged as a common strategy. 
Most existing methods operate in the time-domain, for example by merging adjacent intervals, lowering temporal resolution, or pooling local segments~\cite{liu2022pyraformer,nie2022time}. 
However, local time-domain segments entangle multi-scale information. 
As a result, it does not separate functional information, making reliable information preservation difficult. 

In contrast, the frequency domain provides a more measurable representation of TS, where physical components such as trends, periodic patterns, and disturbances often manifest as interpretable spectral structures~\cite{zhou2022fedformer,liu2022pyraformer}.
As an illustrative baseline, white noise represents an extreme case of spectral dispersion.
For zero-mean white noise, the autocorrelation reduces to a Dirac delta and the corresponding power spectral density (PSD) is constant:
$
S_x(\omega) = \sigma^2 .
$
It implies that signal energy is uniformly distributed over the entire frequency range, rendering the frequency-domain representation non-compressible. Fortunately, real-world TS are rarely purely random and contain structured components, as shown in Fig.~\ref{fig:time_vs_frequency_comparison}. According, a canonical decomposition of a TS can be expressed as:
\begin{equation}
x(t) = m(t) + \sum_{k=1}^{K} a_k \cos(\omega_k t + \phi_k) + \varepsilon(t),
\end{equation}
where \(m(t)\) denotes a slowly varying trend component, \(\sum_{k=1}^{K} a_k \cos(\omega_k t + \phi_k)\) captures periodic patterns, and \(\varepsilon(t)\) represents stochastic disturbances. Correspondingly, the PSD of such signals can be approximated as:
\begin{equation}
S_x(\omega) \approx S_m(\omega) + \sum_{k=1}^{K} c_k \delta(\omega - \omega_k) + S_{\varepsilon}(\omega),
\end{equation}
indicating that the majority of signal energy is concentrated in a small number of  frequency bands. Therefore, the frequency-domain offers a principled space for removing redundancy.

\subsection{TS Token Merging}
In the following, we investigate TS token merging from a frequency-domain perspective.
Given the TS tokens produced by a TS embedder, we denote
$\mathbf{X} = \{ \mathbf{x}_i \}_{i=1}^{N}$ with $\mathbf{x}_i \in \mathbb{R}^{D}$,
where $N$ denotes the number of tokens and $D$ is the feature dimension.
We perform a discrete Fourier transform (DFT) along the token axis,
mapping the token sequence from the time-domain into the frequency-domain as:
$
\mathbf{F} = \{ \mathbf{F}_k \}_{k=0}^{N-1},  \mathbf{F}_k \in \mathbb{C}^{D}.
$

Due to the conjugate symmetry of real-valued TS in the frequency-domain, we retain only the non-negative frequency components $k = 0, \ldots, K-1$, where $K = \lfloor N/2 \rfloor + 1$.
For the $b$-th sample, its spectral magnitude is
$\hat{F}_{b,i,k} = |F_{b,i,k}|$. We sort tokens and select the minimal subset whose cumulative magnitude accounts for at least $95\%$ of the total magnitude at that frequency. Tokens in this subset are treated as the active token set $\mathcal{I}_{b,k}$, and its size is denoted by $n_{b,k} = |\mathcal{I}_{b,k}|$.

The next core question is \emph{how to determine the merging granularity for each frequency component?} Specifically, we denote by $m_{b,k}$ the number of TS tokens after merging, which satisfies $1 \le m_{b,k} \le n_{b,k}$. In practice, selecting an appropriate ratio is difficult. So we introduce an {adaptive frequency merging} strategy. Our key idea is to quantify the structural contribution of each frequency component through its \emph{spectral identifiability}, estimated from locally stationary token segments, where tokens enabling sharper peaks or low-noise periodic patterns contribute more.

Let $I_{b,i}(\omega)$ denote the periodogram of this local segment, and let $S(\omega;\boldsymbol{\theta}_{b,k})$ be a parametric PSD model with parameters $\boldsymbol{\theta}_{b,k}$. Using the Whittle likelihood approximation, the local log-likelihood is defined as:
\begin{equation}
\ell_{b,i}(\boldsymbol{\theta}_{b,k})
=
-\sum_{\omega \in \Omega_k}
\left(
\log S(\omega;\boldsymbol{\theta}_{b,k})
+
\frac{I_{b,i}(\omega)}{S(\omega;\boldsymbol{\theta}_{b,k})}
\right),
\end{equation}
where $\Omega_k$ denotes a frequency neighborhood centered. We measure how strongly token $i$ supports the identification through the Fisher information~\cite{casella2024statistical}:
\begin{equation}
\mathcal{I}_{b,i,k}
=
-\mathbb{E}\!\left[
\nabla^2_{\boldsymbol{\theta}_{b,k}}
\ell_{b,i}(\boldsymbol{\theta}_{b,k})
\right],
\end{equation}
and use its trace $s_{b,i,k} = \mathrm{tr}\!\left( \mathcal{I}_{b,i,k} \right)$ as the contribution. We aggregate contributions into a score:
\begin{equation}
S_{b,k} = \frac{1}{n_{b,k}} \sum_{i \in \mathcal{I}_{b,k}} s_{b,i,k},
\end{equation}

Then, we convert scores into a merging strength $r_{b,k}$ via a sigmoid mapping. The adaptive merging budget is determined as:
$
m_{b,k}
=
\!
\left\lceil (1 - r_{b,k}) \cdot n_{b,k} \right\rceil
.
$
We introduce a normalized regularization term to encourage compact representations while keeping task performance:
$
L_{\mathrm{pen}}
=
\lambda
{\sum_k m_{b,k}}/{\sum_k n_{b,k}}.
$
After determining the merging budget $m_{b,k}$, we first build a sparse local graph by computing similarities only within a neighborhood $|i-j|<r_{\text{loc}}$:
\begin{equation}
\begin{aligned}
\text{sim}_{i,j}
&=
\text{sim}\big(\phi(|F_{i,k}|),\,\phi(|F_{j,k}|)\big), \\
A_{i,j}
&=
\exp\!\left(\text{sim}_{i,j}/\tau\right),
\end{aligned}
\end{equation}
and the affinity matrix $A_{i,j}$ constrains which tokens can be merging. Let $\{\mathcal{G}_{b,k}^{(u)}\}_{u=1}^{m_{b,k}}$ denote a partition of the active indices $\mathcal{I}_{b,k}$ into $m_{b,k}$ local groups. For each group, the merged representative is:
\begin{equation}
\begin{aligned}
\omega_{i}^{(u,k)}
&=
\frac{\sum_{j \in \mathcal{G}^{(u)}_{b,k}} A_{i,j}}
{\sum_{p \in \mathcal{G}^{(u)}_{b,k}} \sum_{q \in \mathcal{G}^{(u)}_{b,k}} A_{p,q}},\\
\tilde{\mathbf{F}}_{u,k}
&=
\sum_{i \in \mathcal{G}^{(u)}_{b,k}}
\omega_{i}^{(u,k)}\, \mathbf{F}_{i,k},
\end{aligned}
\end{equation}

Finally, $\{\tilde{\mathbf{F}}_{u,k}\}_{k=0}^{K-1}$ are transformed back to the time-domain via an IDFT along the frequency axis.

\subsection{Prompt Tokens Compression with Pyramidal Decay}
In typical TS–text tasks, text prompts provide high-level guidance rather than describing numerical variations. As TS representations progressively abstract from low-level patterns to higher-level semantics, textual guidance plays a stronger role in shallow layers but diminishes with depth, where representation learning increasingly relies on the TS itself. Consequently, retaining full prompt tokens across all layers offers limited benefit while incurring additional overhead, suggesting that prompt influence should be gradually absorbed into TS representations at early stages.

Let the prompt tokens at layer $l$ be
$\mathbf{H}_P^{(l)} \in \mathbb{R}^{P_l \times D}$, where the number of prompt
tokens is reduced as
$P_{l+1} = \lceil 0.25\, P_l \rceil$.
We first compress the prompt tokens
$\mathbf{S}^{(l)} = \mathbf{W}^{(l)} \mathbf{H}_P^{(l)}$,
with $\mathbf{W}^{(l)} \in \mathbb{R}^{P_{l+1} \times P_l}$ whose rows are
normalized by a softmax.
To preserve global prompt information, we additionally compute a residual
summary $\mathbf{r}^{(l)} = \mathrm{MeanPool}(\mathbf{H}_P^{(l)})$.
The prompt tokens for the next layer is then updated as:
\begin{equation}
\mathbf{H}_P^{(l+1)} = \mathbf{S}^{(l)} + \mathbf{1}_{P_{l+1}}\, \mathbf{r}^{(l)} .
\label{eq:prompt-compression}
\end{equation}

As depth increases, the number of prompt tokens decays rapidly (25\%, 6.3\%, 1.6\%, 0.4\%, $\ldots$) relative to the input. Therefore, from around the fourth layer, the computational overhead is negligible.
\section{Experiments}
\subsection{Preparations}
\noindent\textbf{LLM-based TS Models.}\quad We conduct experiments on four advanced models, including OFA~\cite{zhou2023one}, TimeLLM~\cite{jin2023time}, CALF~\cite{liu2025calf} and S\textsuperscript{2}IP~\cite{pan2024s}, referring to App.~\ref{app:LLM-based TS Models}. By spanning a wide range of TS–text alignment strategies for adapting LLMs to time series analysis, this evaluation provides a rigorous and challenging testbed.

\smallskip
\noindent\textbf{TS Analysis Tasks and Datasets.}\quad
Based on the 7 subtasks in Sec.~\ref{TS Tasks: From Patterns to Reasoning}, including \emph{\textbf{Pattern-centric Tasks (Imputation, Classification, Anomaly Detection, and Forecasting)}} and \emph{\textbf{Reasoning-centric Tasks (Simple Deterministic, Complex Deterministic, and Probabilistic)}}, we collect a total of 27 datasets. These datasets span a wide range of domains, including climate, energy, transportation, economics, physiology, chemistry, speech, machine monitoring, cybersecurity, and industrial systems, all sourced from real-world settings. 

\smallskip
\noindent\textbf{Baseline Token Merging Methods.}\quad We include three advanced token merging methods as baselines. \emph{\textbf{Global token merging for TS}}~\cite{shao2026holitom} reduces the number of tokens at each layer by merging the most similar token pairs based on global similarity. \emph{\textbf{Local token merging for TS}}~\cite{bolya2022token} introduces locality constraints by computing token similarity and performing merging only within adjacent or local windows, which preserves temporal ordering and local dependency structures. \emph{\textbf{Dynamic token merging}}~\cite{feng2023efficient} adaptively adjusts the merging ratio according to token similarity, allowing the merging process to flexibly respond to different inputs and model depths. Please refer to App.~\ref{app:Token Merging Methods}.

\subsection{Evaluation on Pattern-centric Tasks}
\label{Evaluation on Pattern-centric Tasks}
\noindent\textbf{Setups.}\quad \textbf{For the imputation task}, we evaluate four models on six datasets. For each dataset, we randomly mask $\{12.5\%, 25\%, 50\%\}$ of time points from TS segments of length 96 to control the missing data ratio, following the protocol of TimesNet. MSE and MAE are adopted as evaluation metrics. This results in 12 comparative experiments per dataset and 72 comparative experiments across 6 datasets, comparing the original models with TokenDecouple (TS token merging + Prompt token compression). \textbf{For the classification task}, we evaluate four models on five datasets using Accuracy as the evaluation metric, resulting in 20 comparative experiments. \textbf{For the anomaly detection task}, four models are evaluated on five datasets with F1-score as the metric. To ensure a fair comparison, only the classical reconstruction error is adopted, consistent with the setting in TimesNet, leading to 20 comparative experiments. \textbf{For the forecasting task}, we adopt two prediction horizons $H \in \{192, 720\}$. The input TS length $T$ is set to 336, 512, 96, and 512 for OFA, TimeLLM, CALF, and S\textsuperscript{2}IP, respectively, following their original papers. MSE and MAE are used as evaluation metrics. Each dataset involves 8 experiments, resulting in 40 comparative experiments.

\smallskip
\noindent\textbf{Results.}\quad Across the four tasks, the TokenDecouple models outperform the original models in 76\%, 60\%, 85\%, and 85\% of the cases, respectively, as shown in Tab.~\ref{tab:results1}. Importantly, even in the loss cases, the maximum performance degradation remains minor across all tasks. Specifically, the largest increase in imputation MAE is only \textcolor{black}{0.008} (\textcolor{black}{2.2\%}), the maximum drop in classification accuracy is \textcolor{black}{1.3\%}, the largest decrease in anomaly detection F1-Score is \textcolor{black}{0.41\%}, and for forecasting, the maximum MAE increase is \textcolor{black}{0.01} (\textcolor{black}{5.0\%}).

\begin{table*}[t]
\centering
\footnotesize
\setlength{\tabcolsep}{8pt}
\caption{Performance comparison between the original models (Ori) and TokenDecouple (TD) on pattern-centric tasks. Across 152 comparative experiments, TokenDecouple achieves better performance in 118 cases, accounting for 78\%, while significantly reducing token-level computational cost. In 1.3\% of the cases, the performance remains on par with the original models. For the Imputation task, MAE is averaged over three masking ratios. For the Forecasting task, MAE is averaged over two prediction horizons.}
\begin{adjustbox}{max width=\textwidth}
\begin{tabular}{c|cc|cc|cc|cc|ccc|c|c} % 
\toprule \toprule
\multirow{2}{*}{\textbf{Models}} & \multicolumn{2}{c|}{\textbf{OFA}} &  \multicolumn{2}{c|}{\textbf{TimeLLM}} & \multicolumn{2}{c|}{\textbf{CALF}} &  \multicolumn{2}{c|}{\textbf{S\textsuperscript{2}IP}}& \multirow{2}{*}{\textbf{Wins}}  & \multirow{2}{*}{\textbf{Draws}} & \multirow{2}{*}{\textbf{Losses}} & \multirow{2}{*}{\textbf{Total}}& \multirow{2}{*}{\textbf{Win Rate$\uparrow$}}\\ 
\cmidrule(lr){2-3} \cmidrule(lr){4-5} \cmidrule(lr){6-7} \cmidrule(lr){8-9}
& Ori & TD & Ori & TD& Ori & TD& Ori & TD &&&&&\\
% \cmidrule(lr){2-3} \cmidrule(lr){4-5} \cmidrule(lr){6-7} \cmidrule(lr){8-9}
\midrule \midrule
\multicolumn{14}{c}{\textbf{Imputation Task (Metric: avg. MAE$\downarrow$)}} \\ 
\cdashline{1-14}
\noalign{\vskip 1.0mm}
ETTh1 &0.166&0.162&0.089&0.087&0.302&0.297&0.135&0.126&11&0&1&12&92\%\\
ETTh2 &0.131&0.128&0.141&0.138&0.378&0.363&0.104&0.106&9&0&3&12&75\%\\
ETTm1 &0.097&0.095&0.071&0.073&0.306&0.300&0.092&0.091&9&0&3&12&75\%\\
ETTm2 &0.079&0.074&0.069&0.067&0.285&0.279&0.043&0.044&8&1&3&12&67\%\\
Weather &0.050&0.047&0.043&0.040&0.122&0.120&0.044&0.040&10&0&2&12&83\%\\
Electricity &0.197&0.197&0.079&0.078&0.237&0.225&0.187&0.178&8&1&3&12&67\%\\
\noalign{\vskip 1.0mm}
\cdashline{1-14}
\noalign{\vskip 1.0mm}
\multicolumn{14}{c}{Wins: 55, Draws: 2, Losses: 15, average Win Rate: 76\%} \\
\midrule \midrule
\multicolumn{14}{c}{\textbf{Classification Task (Metric: Accuracy$\uparrow$ (\%))}} \\ 
\cdashline{1-14}
\noalign{\vskip 1.0mm}
JapaneseVowels & 98.2 & 98.2& 30.1 & 31.2 & 95.9 & 95.3 & 93.6 & 93.8 &2 & 1 & 1 & 4 & 50\% \\
Handwriting & 35.2 & 35.9 & 12.5& 12.3 & 27.1 & 28.9 & 27.7 & 30.4 & 3 & 0 & 1 & 4 & 75\% \\
Heartbeat & 76.8 & 77.1 & 72.9 & 72.4 & 78.7 & 77.4 & 72.1& 72.7 & 2 & 0 & 2 & 4 & 50\% \\
EthanolConcentration & 34.8 & 35.0 & 31.2 & 32.9 & 29.6 & 29.8 & 29.7 & 29.4 & 3 & 0 & 1 & 4 & 75\% \\
PEMS-SF & 84.2 & 85.5 & 25.1 & 27.8 & 78.9 & 78.5 & 82.8 & 82.5 & 2 & 0 & 2 & 4 & 50\% \\
\noalign{\vskip 1.0mm}
\cdashline{1-14}
\noalign{\vskip 1.0mm}
\multicolumn{14}{c}{Wins: 12, Draws: 1, Losses: 7, average Win Rate: 60\%} \\
\midrule \midrule
\multicolumn{14}{c}{\textbf{Anomaly Detection Task (Metric: F1-Score$\uparrow$ (\%))}} \\ 
\cdashline{1-14}
\noalign{\vskip 1.0mm}
SMD & 85.74 & 85.79 & 84.12 & 84.60 & 86.54 & 87.08 & 83.34 & 83.55 & 4 & 0 & 0 & 4 & 100\% \\
MSL & 85.11 & 85.68 & 69.76 & 70.12 & 84.68 & 84.62 & 80.22 & 80.96 & 3 & 0 & 1 & 4 & 75\% \\
SMAP & 70.17 & 72.32 & 70.44 & 71.18 & 68.60 & 68.87 & 71.85 & 72.26 & 4 & 0 & 0 & 4 & 100\% \\
SWaT & 93.09 & 92.86 & 91.63 & 92.48 & 94.56 & 94.91 & 83.46 & 85.11 & 3 & 0 & 1 & 4 & 75\%  \\
PSM & 97.76 & 97.35 & 92.46 & 93.33 & 96.24 & 96.89 & 95.61 & 96.54 & 3 & 0 & 1 & 4 & 75\%  \\
\noalign{\vskip 1.0mm}
\cdashline{1-14}
\noalign{\vskip 1.0mm}
\multicolumn{14}{c}{Wins: 17, Draws: 0, Losses: 3, average Win Rate: 85\%} \\
\midrule \midrule
\multicolumn{14}{c}{\textbf{Forecasting Task (Metric: avg. MAE$\downarrow$)}} \\ 
\cdashline{1-14}
\noalign{\vskip 1.0mm}
Traffic & 0.416 & 0.404 & 0.402 & 0.397 & 0.461 & 0.458 & 0.404 & 0.405 & 7 & 0 & 1 & 8 & 88\% \\
Solar   & 0.198 & 0.204 & 0.175 & 0.166 & 0.250 & 0.248 & 0.191 & 0.186 & 5 & 0 & 3 & 8 & 63\% \\
PEMS04  & 0.513 & 0.509 & 0.486 & 0.485 & 0.779 & 0.770 & 0.581 & 0.577 & 6 & 0 & 2 & 8 & 75\% \\
PEMS08  & 0.566 & 0.548 & 0.557 & 0.538 & 0.859 & 0.855 & 0.734 & 0.725 & 8 & 0 & 0 & 8 & 100\% \\
Exchange &0.511 & 0.490 & 0.553 & 0.538 & 0.503 & 0.486 & 0.599 & 0.590 & 8 & 0 & 0 & 8 & 100\% \\
\noalign{\vskip 1.0mm}
\cdashline{1-14}
\noalign{\vskip 1.0mm}
\multicolumn{14}{c}{Wins: 34, Draws: 0, Losses: 6, average Win Rate: 85\%} \\
\bottomrule \bottomrule
\end{tabular}
\end{adjustbox}
\label{tab:results1}
\end{table*}

\begin{table*}[t]
\centering
\footnotesize
\setlength{\tabcolsep}{8pt}
\caption{Performance comparison between the original models (Ori) and TokenDecouple (TD) on reasoning-centric tasks, evaluated using accuracy(\%). Across 24 comparative experiments, TokenDecouple achieves better performance in 17 cases, accounting for 71\%.}
\begin{adjustbox}{max width=\textwidth}
\begin{tabular}{c|cc|cc|cc|cc|cc|cc} % 
\toprule \toprule
\multirow{3}{*}{\textbf{Tasks}} & \multicolumn{4}{c|}{\textbf{Simple Deterministic}} &  \multicolumn{4}{c|}{\textbf{Complex Deterministic}} & \multicolumn{4}{c}{\textbf{Probabilistic}}\\ 
\cmidrule(lr){2-5} \cmidrule(lr){6-9} \cmidrule(lr){10-13}
& \multicolumn{2}{c}{\textbf{RCW}} & \multicolumn{2}{c|}{\textbf{TEE}} & \multicolumn{2}{c}{\textbf{ECG}} & \multicolumn{2}{c|}{\textbf{EMG}} & \multicolumn{2}{c}{\textbf{CTU}} & \multicolumn{2}{c}{\textbf{HAR}} \\
& Ori & TD & Ori & TD& Ori & TD& Ori & TD & Ori & TD& Ori & TD\\
\midrule \midrule
OFA & 78.20 & 80.25 & 62.31 & 64.39 & 21.39 & 20.44 & 75.45 & 75.27  & 69.88 & 71.62& 87.64 & 88.25 \\
\midrule
TimeLLM & 62.50 & 64.53 & 47.28 & 50.36 & 25.59 & 23.68 & 63.84 & 65.26 & 60.20 & 61.84& 88.43 & 85.22 \\
\midrule
CALF & 74.56 & 77.48 & 53.44 & 57.86 & 24.24 &  23.55 & 66.73 & 69.31 & 55.62 &56.08 & 84.97 & 86.60 \\
\midrule
S\textsuperscript{2}IP & 81.50 & 78.43 & 59.16 & 64.28 & 27.45 & 30.54  & 82.68 &  80.27 & 64.19 & 66.39& 75.22 &  77.44\\
\bottomrule \bottomrule
\end{tabular}
\end{adjustbox}
\label{tab:results2}
\end{table*}

\subsection{Evaluation on Reasoning-centric Tasks}
\noindent\textbf{Setups.}\quad 
The reasoning-centric tasks cover three levels of complexity. Simple Deterministic Reasoning includes Right Whale Call detection (RCW) and Transient Electromagnetic Event classification (TEE), where class labels correspond to well-defined single patterns identified by human annotators. Complex Deterministic Reasoning involves ECG record diagnosis and EMG signal diagnosis, requiring label assignment based on the joint interpretation of multiple temporal patterns by medical experts. Probabilistic Reasoning comprises Human Activity Recognition (HAR) and Computer Type Usage classification (CTU), in which class boundaries are inherently uncertain due to stochastic feature–label relationships.

\smallskip
\noindent\textbf{Results.}\quad We compare the performance of the original models and TokenDecouple across four models and six subtasks, as shown in Tab.~\ref{tab:results2}. Among the 24 task settings, TokenDecouple achieves better performance in 17 cases, accounting for 71\% . Notably, even in the few cases with performance drops, the maximum accuracy degradation is only 3.2\%.

\subsection{Efficiency Evaluation of TokenDecouple}
\begin{figure}[t]
  \includegraphics[width=\columnwidth]{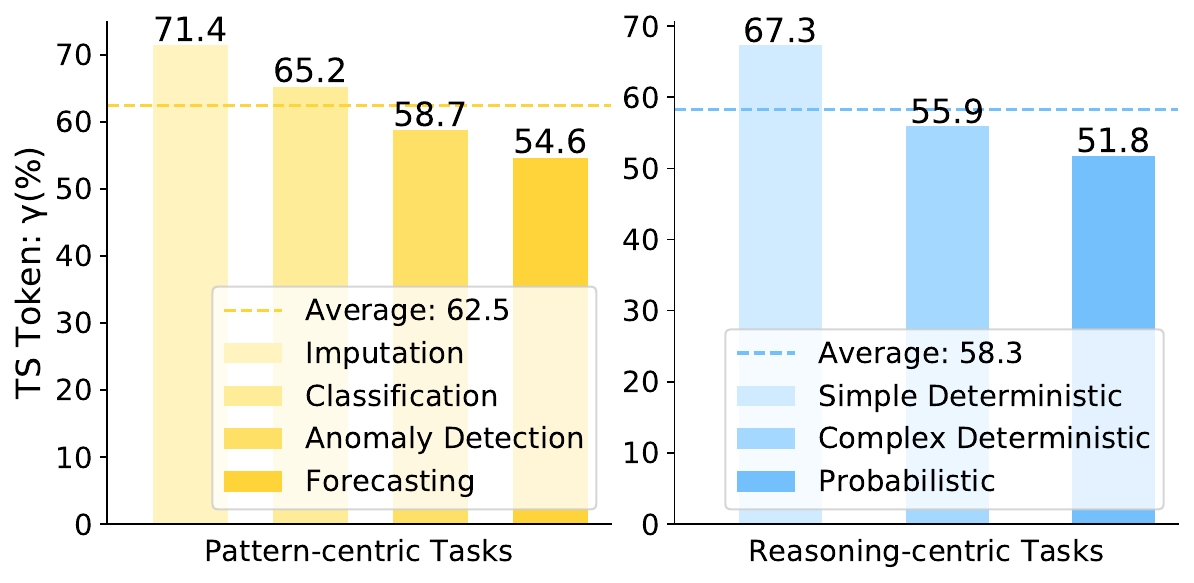}
  \caption{Average TS token merging ratio $\gamma$ (\%), induced by frequency-domain merging across seven tasks.}
  \label{fig:ts_token_compression}
\end{figure}

\noindent\textbf{Computational Cost of TS Tokens.}\quad Given $N_{\text{TS}}$ TS tokens, token merging reduces them to $\hat{N}_{\text{TS}}$ tokens. 
We define the merging ratio as
$\gamma = 1 - \hat{N}_{\text{TS}} / N_{\text{TS}}$,
where a larger $\gamma$ indicates a higher degree of merging.
Fig.~\ref{fig:ts_token_compression} shows the average $\gamma$ of four models across seven tasks.
As task difficulty increases, $\gamma$ gradually decreases, reflecting the growing amount of information required.
Nevertheless, even for the most challenging tasks, at least 50\% the TS tokens can still be merged.

\noindent\textbf{Computational Cost of Prompt Tokens.}\quad In addition, given $N_{\text{Text}}$ prompt tokens, their number follows a pyramidal decay across layers, taking values of 100\%, 25\%, 6.3\%, 1.6\%, and 0.4\%.
Specifically, the number of prompt tokens at layer $l$ is
$N_{\text{Text}}^{(l)} = 0.25^{\,l} N_{\text{Text}}$.
The total number of prompt tokens processed across $L$ layers is
$\sum_{l=1}^{L} N_{\text{Text}}^{(l)} = N_{\text{Text}} (1 - 0.25^{L}) / (1 - 0.25)$,
leading to an average per layer of
$\bar{N}_{\text{Text}} = \frac{N_{\text{Text}}}{L(1-0.25)} (1 - 0.25^{L})$.
When $L \ge 4$, the term $0.25^{L}$ is negligible, yielding
$\bar{N}_{\text{Text}} \approx \frac{4}{3L} N_{\text{Text}}$.
Accordingly, for an $L$-layer LLM, the prompt token compression ratio is
$\eta_{\text{L}} = 1 - \bar{N}_{\text{Text}} / N_{\text{Text}} \approx 1 - \frac{4}{3L}$.
Consequently, prompt-induced overhead decreases sharply with model depth,
with $\eta_{12} \approx 89\%$ and $\eta_{24} \approx 94\%$ for 12-layer and 24-layer models, respectively.

\begin{figure}[t]
  \includegraphics[width=\columnwidth]{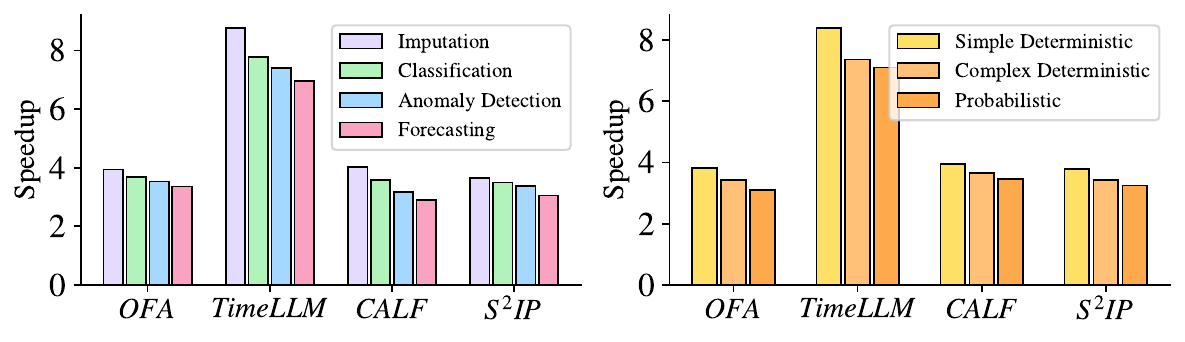}
  \caption{Inference speedup of four models with TokenDecouple across seven tasks.}
  \label{fig:speedup_2subplots}
  \vspace{-0.4cm}
\end{figure}

\smallskip
\noindent\textbf{Inference Speedup.}\quad Under identical experimental settings, we evaluate the inference-time acceleration brought by TokenDecouple, as illustrated in Fig.~\ref{fig:speedup_2subplots}. Consistent inference speedups are observed across seven tasks, with OFA, TimeLLM, CALF, and S\textsuperscript{2}IP achieving average speedups of \textbf{3.55×}, \textbf{7.68×}, \textbf{3.53×}, and \textbf{3.43×}, respectively. TS merging ratios positively correlate with inference speedups, while model architectures further influence the overall acceleration.

\begin{figure}[h]
  \includegraphics[width=\columnwidth]{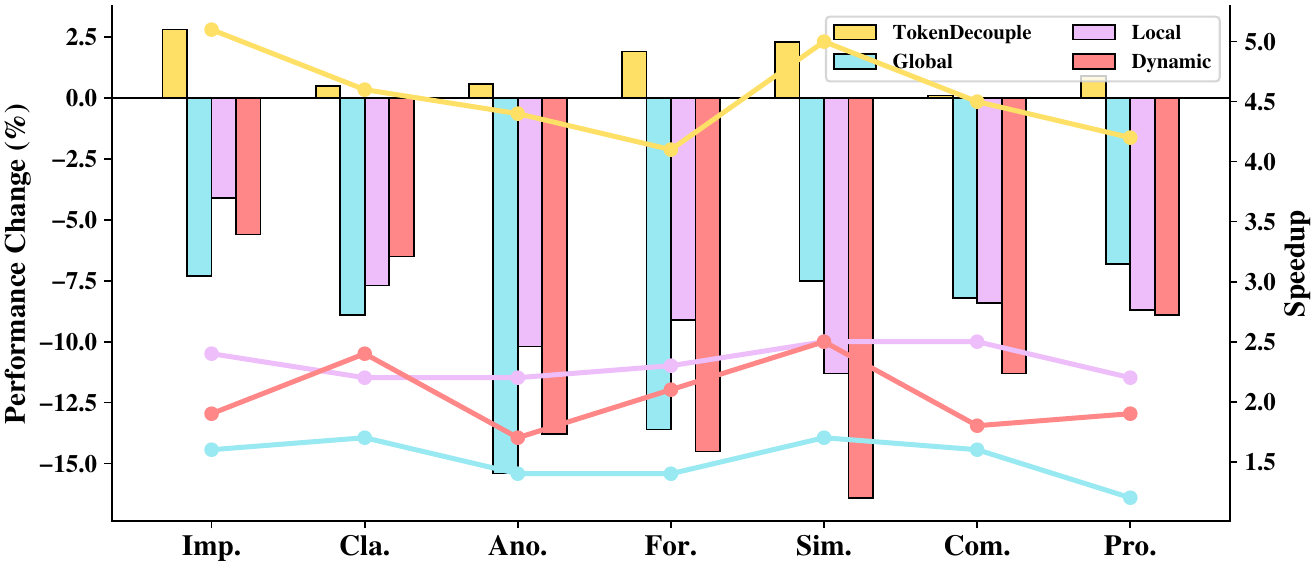}
  \caption{Performance change (bar chart, “+” indicates improvement and “-” indicates degradation) and inference speedup (line chart) of TokenDecouple and three token merging methods, averaged over four models across seven tasks.}
  \label{fig:perf_speedup_combined}
  \vspace{-0.4cm}
\end{figure}

\subsection{Comparison with Other Methods} 
Fig.~\ref{fig:perf_speedup_combined} compares TokenDecouple with three representative token merging methods, namely global merging, local merging, and dynamic merging, in terms of both performance and inference speedup. While all three baselines exhibit noticeable performance degradation across tasks, TokenDecouple consistently preserves competitive performance. Moreover, TokenDecouple achieves an average inference speedup of \textbf{4.55×}, substantially surpassing global, local, and dynamic merging, which obtain average speedups of 1.2×, 2.2×, and 1.9×, respectively. These results indicate that TokenDecouple effectively balances efficiency and performance.

\begin{figure}[t]
  \includegraphics[width=\columnwidth]{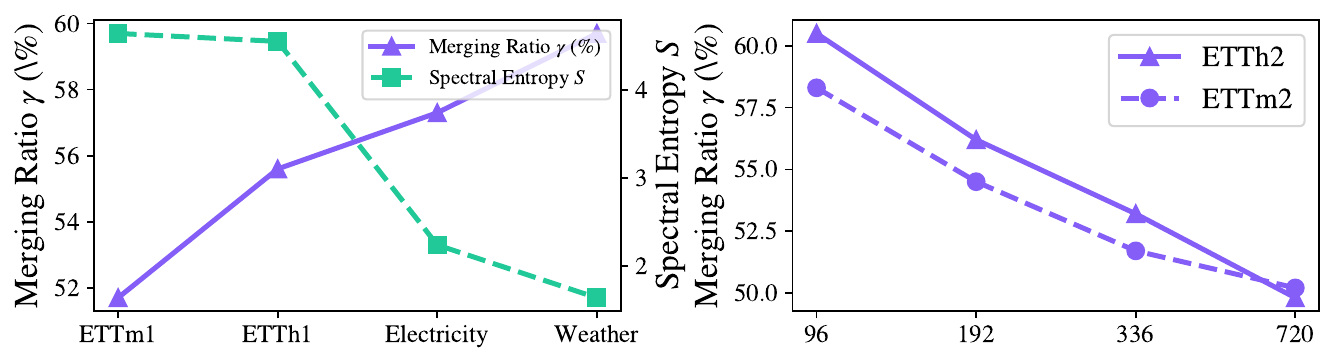}
  \caption{Merging ratio $\gamma$ under different datasets or prediction horizons, averaged over four models.}
  \label{fig:gamma_entropy_and_predlen}
\end{figure}
\subsection{Ablation Study}
Based on the ablation analysis shown in Tab.~\ref{Ablation}, we verify that incorporating text prompts benefits LLM-based TS analysis systems (A vs. B). Moreover, A vs. C and A vs. D show that prompt token compression and TS token merging each introduce only limited performance degradation when applied individually. When combined, they achieve a better trade-off between cost and performance.
\begin{table}[t]
\centering
\caption{
Ablation analysis.
\textbf{A}: Original models;
\textbf{B}: A - text prompt;
\textbf{C}: A + prompt token compression;
\textbf{D}: A + TS token merging;
\textbf{E}: A + prompt token compression + TS token merging.
\{\textbf{Imp.}, \textbf{For.}\} metric: avg. MSE over four models (lower is better);
\{\textbf{Ano.}\} metric: avg. F1-score (\%) over four models (higher is better);
\{\textbf{Cla.}, \textbf{Sim.}, \textbf{Com.}, \textbf{Pro.}\} metric: avg. accuracy (\%) over four models (higher is better).
}

\resizebox{1\linewidth}{!}{
\begin{tabular}{c|ccccccc}
\toprule \toprule
\textbf{Types} & \textbf{Imp.} & \textbf{Cla.}& \textbf{Ano.}& \textbf{For.}& \textbf{Sim.}& \textbf{Com.}& \textbf{Pro.} \\
\midrule
A & 0.144 & 55.86 & 84.27 & 0.487 & 64.87 & 48.42 & 73.26 \\
B & 0.152 & 54.15 & 82.16 & 0.513 & 58.22 & 41.39 & 62.39 \\
\noalign{\vskip 1.0mm}
\cdashline{1-8}
\noalign{\vskip 1.0mm}
C & 0.144 & 55.60 & 83.96 & 0.495 & 64.55 & 48.30 & 72.91 \\
D & 0.141 & 56.42 & 84.79 & 0.477 & 67.11 & 48.46 & 73.98 \\
\noalign{\vskip 1.0mm}
\cdashline{1-8}
\noalign{\vskip 1.0mm}
E & 0.140 & 56.35 & 84.83 & 0.479 & 67.20 & 48.54 & 74.18 \\
\bottomrule \bottomrule
\end{tabular}}
% \vspace{-0.2cm}
\label{Ablation}
\end{table}
\begin{table}[t]
\centering
\caption{MAE of Original (Ori) \& TokenDecouple (TD) under different SNR levels (dB) on forecasting tasks with a prediction horizon of 196.}
\resizebox{1\linewidth}{!}{
\begin{tabular}{c|c|ccc|ccc}
\toprule \toprule
\multicolumn{2}{c|}{\textbf{Models}} & \multicolumn{3}{c|}{OFA} & \multicolumn{3}{c}{TimeLLM} \\
\midrule
\multicolumn{2}{c|}{\textbf{SNR(dB)}} & \textbf{10}& \textbf{20}& \textbf{30}& \textbf{10}& \textbf{20}& \textbf{30}\\
\midrule
\multirow{2}{*}{\textbf{Traffic}} & Ori &0.438 &0.423 & 0.408&0.424 &0.405 &0.381\\
& TD &0.417 &0.411 & 0.404&0.410 & 0.399& 0.383\\
\midrule
\multirow{2}{*}{\textbf{Exchange}} & Ori &0.258 & 0.218&0.192 &0.277 &0.233 &0.215\\
& TD & 0.230&0.214 & 0.196& 0.256&0.228 & 0.211\\
\bottomrule\bottomrule
\end{tabular}}
\label{Robustness}
\end{table}

\section{Discussion \& Conclusion}
\noindent\textbf{TS Token Merging Ratio Decreases with Increasing Task and Dataset Difficulty.}\quad As shown in Fig.~\ref{fig:ts_token_compression}, the ratio $\gamma$ exhibits a clear decreasing trend across tasks of increasing difficulty. Furthermore, focusing on the forecasting task, we observe that $\gamma$ also decreases as the \textbf{prediction horizon} becomes longer or as the dataset distribution exhibits higher \textbf{spectral entropy}. This indicates more TS tokens is required under higher prediction difficulty or increased data complexity, as shown in Fig.~\ref{fig:gamma_entropy_and_predlen}.

\smallskip
\noindent\textbf{Robustness Analysis of TokenDecouple.}\quad Taking the forecasting on the Weather and Exchange as examples, we inject additive Gaussian white noise with SNR $\in \{30, 20, 10\}$ dB to examine whether TokenDecouple exhibits failure. Tab.~\ref{Robustness} reports the performance of OFA and TimeLLM under different SNR levels, both with and without TokenDecouple. We observe that as the noise level increases (i.e., SNR decreases), the $|\mathrm{MAE}_{\text{Ori}} - \mathrm{MAE}_{\text{TD}}|$ enlarges. This indicates that token merging enhances robustness to noise-induced frequency perturbations.

\smallskip
\noindent\textbf{Conclusion.}\quad
Our work revisits LLM-based TS analysis from a token-centric perspective. We show that TS tokens contain frequency-domain redundancy, while prompt tokens exhibit a pyramidal decay effect. Based on these findings, we propose \textbf{\textit{TokenDecouple}}, which decouples TS token merging and prompt token compression. Extensive experiments show that TokenDecouple improves inference efficiency while maintaining or enhancing performance, highlighting the value of token-aware design for efficient TS--language integration.

\section*{Limitations}
Although TokenDecouple demonstrates a favorable efficiency–performance trade-off across multiple time series tasks, this study does not systematically evaluate its applicability or robustness under highly non-stationary or irregularly sampled conditions, which are common in certain real-world scenarios. Our experiments primarily focus on general time series settings, and the behavior of the proposed token decoupling strategy in more extreme, noisy, or rapidly evolving environments remains unclear.
In addition, our study is limited to token-level compression during inference and does not examine its interaction with other efficiency-oriented techniques, such as structured pruning, quantization, or model-level compression and acceleration. As a result, the effectiveness of TokenDecouple in extremely large-scale models or strictly resource-constrained settings has not been fully assessed.
\bibliography{custom}

@inproceedings{roy2024exploring,
  title={Exploring llm-based agents for root cause analysis},
  author={Roy, Devjeet and Zhang, Xuchao and Bhave, Rashi and Bansal, Chetan and Las-Casas, Pedro and Fonseca, Rodrigo and Rajmohan, Saravan},
  booktitle={Companion proceedings of the 32nd ACM international conference on the foundations of software engineering},
  pages={208--219},
  year={2024}
}

@article{qin2025largelanguagemodelsmeet,
  title={Large language models meet nlp: A survey},
  author={Qin, Libo and Chen, Qiguang and Feng, Xiachong and Wu, Yang and Zhang, Yongheng and Li, Yinghui and Li, Min and Che, Wanxiang and Yu, Philip S},
  journal={Frontiers of Computer Science},
  volume={20},
  number={11},
  pages={2011361},
  year={2026},
  publisher={Springer}
}

@article{matarazzo2025surveylargelanguagemodels,
  title={A survey on large language models with some insights on their capabilities and limitations},
  author={Matarazzo, Andrea and Torlone, Riccardo},
  journal={arXiv preprint arXiv:2501.04040},
  year={2025}
}

@article{guo2025aligned,
  title={Aligned better, listen better for audio-visual large language models},
  author={Guo, Yuxin and Ma, Shuailei and Ma, Shijie and Bao, Xiaoyi and Xie, Chen-Wei and Zheng, Kecheng and Weng, Tingyu and Sun, Siyang and Zheng, Yun and Zou, Wei},
  journal={arXiv preprint arXiv:2504.02061},
  year={2025}
}

@inproceedings{zhang2025llava,
  title={Llava-mini: Efficient image and video large multimodal models with one vision token},
  author={Zhang, Shaolei and Fang, Qingkai and Yang, Yang and Feng, Yang},
  booktitle={International Conference on Learning Representations},
  volume={2025},
  pages={53285--53310},
  year={2025}
}

@inproceedings{han2024multimodal,
  title={Multimodal large language models and tunings: Vision, language, sensors, audio, and beyond},
  author={Han, Soyeon Caren and Cao, Feiqi and Poon, Josiah and Navigli, Roberto},
  booktitle={Proceedings of the 32nd ACM International Conference on Multimedia},
  pages={11294--11295},
  year={2024}
}

@article{chang2025llm4ts,
  title={Llm4ts: Aligning pre-trained llms as data-efficient time-series forecasters},
  author={Chang, Ching and Wang, Wei-Yao and Peng, Wen-Chih and Chen, Tien-Fu},
  journal={ACM Transactions on Intelligent Systems and Technology},
  volume={16},
  number={3},
  pages={1--20},
  year={2025},
  publisher={ACM New York, NY}
}

@inproceedings{pan2024s,
  title={S2IP-LLM: Semantic space informed prompt learning with LLM for time series forecasting},
  author={Pan, Zijie and Jiang, Yushan and Garg, Sahil and Schneider, Anderson and Nevmyvaka, Yuriy and Song, Dongjin},
  booktitle={Forty-first International Conference on Machine Learning},
  year={2024}
}

@article{wang2024news,
  title={From news to forecast: Integrating event analysis in llm-based time series forecasting with reflection},
  author={Wang, Xinlei and Feng, Maike and Qiu, Jing and Gu, Jinjin and Zhao, Junhua},
  journal={Advances in Neural Information Processing Systems},
  volume={37},
  pages={58118--58153},
  year={2024}
}

@inproceedings{kong2025time,
  title={Time-mqa: Time series multi-task question answering with context enhancement},
  author={Kong, Yaxuan and Yang, Yiyuan and Hwang, Yoontae and Du, Wenjie and Zohren, Stefan and Wang, Zhangyang and Jin, Ming and Wen, Qingsong},
  booktitle={Proceedings of the 63rd Annual Meeting of the Association for Computational Linguistics (Volume 1: Long Papers)},
  pages={29736--29753},
  year={2025}
}

@inproceedings{wang2025itformer,
  title={ITFormer: Bridging Time Series and Natural Language for Multi-Modal QA with Large-Scale Multitask Dataset},
  author={Wang, Yilin and Lei, Peixuan and Song, Jie and Hao, Yuzhe and Chen, Tao and Zhang, Yuxuan and Jia, Lei and Li, Yuanxiang and Wei, Zhongyu},
  booktitle={International Conference on Machine Learning},
  pages={63324--63344},
  year={2025},
  organization={PMLR}
}

@inproceedings{meunier2025crisists,
  title={CrisisTS: Coupling Social Media Textual Data and Meteorological Time Series for Urgency Classification},
  author={Meunier, Romain and Benamara, Farah and Moriceau, V{\'e}ronique and Qiao, Zhongzheng and Ramasamy, Savitha},
  booktitle={Proceedings of the 63rd Annual Meeting of the Association for Computational Linguistics (Volume 1: Long Papers)},
  pages={16082--16099},
  year={2025}
}

@article{he2024robust,
  title={Robust Multivariate Time Series Forecasting against Intra-and Inter-Series Transitional Shift},
  author={He, Hui and Zhang, Qi and Yi, Kun and Xue, Xiaojun and Wang, Shoujin and Hu, Liang and Cao, Longbing},
  journal={arXiv preprint arXiv:2407.13194},
  year={2024}
}

@inproceedings{lu2025nonstationarytimeseriesforecasting,
  title={Towards non-stationary time series forecasting with temporal stabilization and frequency differencing},
  author={Lu, Junkai and Chen, Peng and Guo, Chenjuan and Shu, Yang and Wang, Meng and Yang, Bin},
  booktitle={Proceedings of the AAAI Conference on Artificial Intelligence},
  volume={40},
  number={29},
  pages={24070--24078},
  year={2026}
}

@book{casella2024statistical,
  title={Statistical inference},
  author={Casella, George and Berger, Roger},
  year={2024},
  publisher={Chapman and Hall/CRC}
}

@inproceedings{zhou2022fedformer,
  title={Fedformer: Frequency enhanced decomposed transformer for long-term series forecasting},
  author={Zhou, Tian and Ma, Ziqing and Wen, Qingsong and Wang, Xue and Sun, Liang and Jin, Rong},
  booktitle={International conference on machine learning},
  pages={27268--27286},
  year={2022},
  organization={PMLR}
}

@inproceedings{jin2023time,
  title={Time-llm: Time series forecasting by reprogramming large language models},
  author={Jin, Ming and Wang, Shiyu and Ma, Lintao and Chu, Zhixuan and Zhang, James and Shi, Xiaoming and Chen, Pin-Yu and Liang, Yuxuan and Li, Yuan-Fang and Pan, Shirui and others},
  booktitle={International conference on learning representations},
  volume={2024},
  pages={23857--23880},
  year={2024}
}

@inproceedings{kwon2023efficient,
  title={Efficient memory management for large language model serving with pagedattention},
  author={Kwon, Woosuk and Li, Zhuohan and Zhuang, Siyuan and Sheng, Ying and Zheng, Lianmin and Yu, Cody Hao and Gonzalez, Joseph and Zhang, Hao and Stoica, Ion},
  booktitle={Proceedings of the 29th symposium on operating systems principles},
  pages={611--626},
  year={2023}
}

@article{dao2022flashattention,
  title={Flashattention: Fast and memory-efficient exact attention with io-awareness},
  author={Dao, Tri and Fu, Dan and Ermon, Stefano and Rudra, Atri and R{\'e}, Christopher},
  journal={Advances in neural information processing systems},
  volume={35},
  pages={16344--16359},
  year={2022}
}

@article{xie2024chatts,
  title={ChatTS: Aligning Time Series with LLMs via Synthetic Data for Enhanced Understanding and Reasoning},
  author={Xie, Zhe and Li, Zeyan and He, Xiao and Xu, Longlong and Wen, Xidao and Zhang, Tieying and Chen, Jianjun and Shi, Rui and Pei, Dan},
  journal={Proceedings of the VLDB Endowment},
  volume={18},
  number={8},
  pages={2385--2398},
  year={2025},
  publisher={VLDB Endowment}
}

@inproceedings{liu2025timecma,
  title={Timecma: Towards llm-empowered multivariate time series forecasting via cross-modality alignment},
  author={Liu, Chenxi and Xu, Qianxiong and Miao, Hao and Yang, Sun and Zhang, Lingzheng and Long, Cheng and Li, Ziyue and Zhao, Rui},
  booktitle={Proceedings of the AAAI Conference on Artificial Intelligence},
  volume={39},
  number={18},
  pages={18780--18788},
  year={2025}
}

@inproceedings{liu2025calf,
  title={Calf: Aligning llms for time series forecasting via cross-modal fine-tuning},
  author={Liu, Peiyuan and Guo, Hang and Dai, Tao and Li, Naiqi and Bao, Jigang and Ren, Xudong and Jiang, Yong and Xia, Shu-Tao},
  booktitle={Proceedings of the AAAI Conference on Artificial Intelligence},
  volume={39},
  number={18},
  pages={18915--18923},
  year={2025}
}

@article{gruver2023large,
  title={Large language models are zero-shot time series forecasters},
  author={Gruver, Nate and Finzi, Marc and Qiu, Shikai and Wilson, Andrew G},
  journal={Advances in Neural Information Processing Systems},
  volume={36},
  pages={19622--19635},
  year={2023}
}

@inproceedings{liu2024lstprompt,
  title={Lstprompt: Large language models as zero-shot time series forecasters by long-short-term prompting},
  author={Liu, Haoxin and Zhao, Zhiyuan and Wang, Jindong and Kamarthi, Harshavardhan and Prakash, B Aditya},
  booktitle={Findings of the Association for Computational Linguistics: ACL 2024},
  pages={7832--7840},
  year={2024}
}

@inproceedings{zhong2025time,
  title={Time-VLM: Exploring Multimodal Vision-Language Models for Augmented Time Series Forecasting},
  author={Zhong, Siru and Ruan, Weilin and Jin, Ming and Li, Huan and Wen, Qingsong and Liang, Yuxuan},
  booktitle={International Conference on Machine Learning},
  pages={78478--78497},
  year={2025},
  organization={PMLR}
}

@article{ruan2025vision,
  title={Vision-Enhanced Time Series Forecasting via Latent Diffusion Models},
  author={Ruan, Weilin and Zhong, Siru and Wen, Haomin and Liang, Yuxuan},
  journal={arXiv preprint arXiv:2502.14887},
  year={2025}
}

@inproceedings{jia2024gpt4mts,
  title={Gpt4mts: Prompt-based large language model for multimodal time-series forecasting},
  author={Jia, Furong and Wang, Kevin and Zheng, Yixiang and Cao, Defu and Liu, Yan},
  booktitle={Proceedings of the AAAI Conference on Artificial Intelligence},
  volume={38},
  number={21},
  pages={23343--23351},
  year={2024}
}

@article{liu2024time,
  title={Time-mmd: Multi-domain multimodal dataset for time series analysis},
  author={Liu, Haoxin and Xu, Shangqing and Zhao, Zhiyuan and Kong, Lingkai and Prabhakar Kamarthi, Harshavardhan and Sasanur, Aditya and Sharma, Megha and Cui, Jiaming and Wen, Qingsong and Zhang, Chao and others},
  journal={Advances in Neural Information Processing Systems},
  volume={37},
  pages={77888--77933},
  year={2024}
}

@article{zhang2025timeseriesanalysisfrequency,
  title={Time series analysis in frequency domain: A survey of open challenges, opportunities and benchmarks},
  author={Zhang, Qianru and Sun, Yuting and Wen, Honggang and Yang, Peng and Li, Xinzhu and Li, Ming and Lam, Kwok-Yan and Yiu, Siu-Ming and Yin, Hongzhi},
  journal={arXiv preprint arXiv:2504.07099},
  year={2025}
}

@inproceedings{yue2025freeformerfrequencyenhancedtransformer,
  title={FreEformer: frequency enhanced transformer for multivariate time series forecasting},
  author={Yue, Wenzhen and Liu, Yong and Ying, Xianghua and Xing, Bowei and Guo, Ruohao and Shi, Ji},
  booktitle={Proceedings of the Thirty-Fourth International Joint Conference on Artificial Intelligence},
  pages={3606--3614},
  year={2025}
}

@article{li2025ftmixer,
  title={FTMixer: Frequency and Time Domain Representations Fusion for Time Series Forecasting},
  author={Li, Zhengnan and Tan, Yuting and Cheng, Xilong and Qin, Yunxiao},
  journal={IEEE Signal Processing Letters},
  year={2025},
  publisher={IEEE}
}

@article{yuan2024d,
  title={D-PAD: Deep-Shallow Multi-Frequency Patterns Disentangling for Time Series Forecasting},
  author={Yuan, Xiaobing and Chen, Ling},
  journal={arXiv preprint arXiv:2403.17814},
  year={2024}
}

@article{he2025unified,
  title={A Unified Frequency Domain Decomposition Framework for Interpretable and Robust Time Series Forecasting},
  author={He, Cheng and Liang, Xijie and Zheng, Zengrong and Lee, Patrick PC and Huang, Xu and Li, Zhaoyi and Xie, Hong and Lian, Defu and Chen, Enhong},
  journal={arXiv preprint arXiv:2510.10145},
  year={2025}
}

@article{qin2025sfdformer,
  title={SFDformer: a frequency-based sparse decomposition transformer for air pollution time series prediction},
  author={Qin, Zhenkai and Wei, Baozhong and Gao, Caifeng and Chen, Xiaolong and Zhang, Hongfeng and In Wong, Cora Un},
  journal={Frontiers in Environmental Science},
  volume={13},
  pages={1549209},
  year={2025},
  publisher={Frontiers Media SA}
}

@inproceedings{
wu2022timesnet,
title={TimesNet: Temporal 2D-Variation Modeling for General Time Series Analysis},
author={Haixu Wu and Tengge Hu and Yong Liu and Hang Zhou and Jianmin Wang and Mingsheng Long},
booktitle={The Eleventh International Conference on Learning Representations },
year={2023},
}

@inproceedings{tang2025llm,
  title={Empowering Large Language Models for Time Series Forecasting with Patterns and Semantics},
  author={Tang, Jialiang and Chen, Shuo and Gong, Chen and Zhang, Jing and Tao, Dacheng},
  booktitle={2025 IEEE International Conference on Data Mining (ICDM)},
  pages={733--742},
  year={2025},
  organization={IEEE}
}

@article{zhou2023one,
  title={One fits all: Power general time series analysis by pretrained lm},
  author={Zhou, Tian and Niu, Peisong and Sun, Liang and Jin, Rong and others},
  journal={Advances in neural information processing systems},
  volume={36},
  pages={43322--43355},
  year={2023}
}

@article{french1999catastrophic,
  title={Catastrophic forgetting in connectionist networks},
  author={French, Robert M},
  journal={Trends in cognitive sciences},
  volume={3},
  number={4},
  pages={128--135},
  year={1999},
  publisher={Elsevier}
}

@inproceedings{liu2022pyraformer,
  title={Pyraformer: Low-complexity pyramidal attention for long-range time series modeling and forecasting},
  author={Liu, Shizhan and Yu, Hang and Liao, Cong and Li, Jianguo and Lin, Weiyao and Liu, Alex X and Dustdar, Schahram},
  booktitle={International conference on learning representations},
year={2022}
}

@inproceedings{nie2022time,
  title={A Time Series is Worth 64 Words: Long-term Forecasting with Transformers},
  author={Nie, Yuqi and Nguyen, Nam H and Sinthong, Phanwadee and Kalagnanam, Jayant},
  booktitle={The Eleventh International Conference on Learning Representations},
  year={2023}
}

@article{shao2026holitom,
  title={HoliTom: Holistic Token Merging for Fast Video Large Language Models},
  author={Shao, Kele and Tao, Keda and Qin, Can and You, Haoxuan and Sui, Yang and Wang, Huan},
  journal={Advances in Neural Information Processing Systems},
  volume={38},
  pages={135547--135570},
  year={2026}
}

@article{bolya2022token,
  title={Token merging: Your vit but faster},
  author={Bolya, Daniel and Fu, Cheng-Yang and Dai, Xiaoliang and Zhang, Peizhao and Feichtenhofer, Christoph and Hoffman, Judy},
  journal={arXiv preprint arXiv:2210.09461},
  year={2022}
}

@article{feng2023efficient,
  title={Efficient vision transformer via token merger},
  author={Feng, Zhanzhou and Zhang, Shiliang},
  journal={IEEE Transactions on Image Processing},
  volume={32},
  pages={4156--4169},
  year={2023},
  publisher={IEEE}
}

@inproceedings{su2019robust,
  title={Robust anomaly detection for multivariate time series through stochastic recurrent neural network},
  author={Su, Ya and Zhao, Youjian and Niu, Chenhao and Liu, Rong and Sun, Wei and Pei, Dan},
  booktitle={Proceedings of the 25th ACM SIGKDD international conference on knowledge discovery \& data mining},
  pages={2828--2837},
  year={2019}
}

@inproceedings{cuturi2011fast,
  title={Fast global alignment kernels},
  author={Cuturi, Marco},
  booktitle={Proceedings of the 28th international conference on machine learning (ICML-11)},
  pages={929--936},
  year={2011}
}

@inproceedings{large2018detecting,
  title={Detecting forged alcohol non-invasively through vibrational spectroscopy and machine learning},
  author={Large, James and Kemsley, E Kate and Wellner, Nikolaus and Goodall, Ian and Bagnall, Anthony},
  booktitle={Pacific-Asia conference on knowledge discovery and data mining},
  pages={298--309},
  year={2018},
  organization={Springer}
}

@article{bagnall2018uea,
  title={The UEA multivariate time series classification archive, 2018},
  author={Bagnall, Anthony and Dau, Hoang Anh and Lines, Jason and Flynn, Michael and Large, James and Bostrom, Aaron and Southam, Paul and Keogh, Eamonn},
  journal={arXiv preprint arXiv:1811.00075},
  year={2018}
}

@article{shokoohi2017generalizing,
  title={Generalizing DTW to the multi-dimensional case requires an adaptive approach},
  author={Shokoohi-Yekta, Mohammad and Hu, Bing and Jin, Hongxia and Wang, Jun and Keogh, Eamonn},
  journal={Data mining and knowledge discovery},
  volume={31},
  number={1},
  pages={1--31},
  year={2017},
  publisher={Springer}
}

@article{kudo1999multidimensional,
  title={Multidimensional curve classification using passing-through regions},
  author={Kudo, Mineichi and Toyama, Jun and Shimbo, Masaru},
  journal={Pattern Recognition Letters},
  volume={20},
  number={11-13},
  pages={1103--1111},
  year={1999},
  publisher={Elsevier}
}

@inproceedings{hundman2018detecting,
  title={Detecting spacecraft anomalies using lstms and nonparametric dynamic thresholding},
  author={Hundman, Kyle and Constantinou, Valentino and Laporte, Christopher and Colwell, Ian and Soderstrom, Tom},
  booktitle={Proceedings of the 24th ACM SIGKDD international conference on knowledge discovery \& data mining},
  pages={387--395},
  year={2018}
}

@inproceedings{mathur2016swat,
  title={SWaT: A water treatment testbed for research and training on ICS security},
  author={Mathur, Aditya P and Tippenhauer, Nils Ole},
  booktitle={2016 international workshop on cyber-physical systems for smart water networks (CySWater)},
  pages={31--36},
  year={2016},
  organization={IEEE}
}

@inproceedings{abdulaal2021practical,
  title={Practical approach to asynchronous multivariate time series anomaly detection and localization},
  author={Abdulaal, Ahmed and Liu, Zhuanghua and Lancewicki, Tomer},
  booktitle={Proceedings of the 27th ACM SIGKDD conference on knowledge discovery \& data mining},
  pages={2485--2494},
  year={2021}
}

@inproceedings{zhou2021informer,
  title={Informer: Beyond efficient transformer for long sequence time-series forecasting},
  author={Zhou, Haoyi and Zhang, Shanghang and Peng, Jieqi and Zhang, Shuai and Li, Jianxin and Xiong, Hui and Zhang, Wancai},
  booktitle={Proceedings of the AAAI conference on artificial intelligence},
  volume={35},
  number={12},
  pages={11106--11115},
  year={2021}
}

@article{li2019enhancing,
  title={Enhancing the locality and breaking the memory bottleneck of transformer on time series forecasting},
  author={Li, Shiyang and Jin, Xiaoyong and Xuan, Yao and Zhou, Xiyou and Chen, Wenhu and Wang, Yu-Xiang and Yan, Xifeng},
  journal={Advances in neural information processing systems},
  volume={32},
  year={2019}
}

@inproceedings{lai2018modeling,
  title={Modeling long-and short-term temporal patterns with deep neural networks},
  author={Lai, Guokun and Chang, Wei-Cheng and Yang, Yiming and Liu, Hanxiao},
  booktitle={The 41st international ACM SIGIR conference on research \& development in information retrieval},
  pages={95--104},
  year={2018}
}

@article{liu2022scinet,
  title={Scinet: Time series modeling and forecasting with sample convolution and interaction},
  author={Liu, Minhao and Zeng, Ailing and Chen, Muxi and Xu, Zhijian and Lai, Qiuxia and Ma, Lingna and Xu, Qiang},
  journal={Advances in neural information processing systems},
  volume={35},
  pages={5816--5828},
  year={2022}
}

@inproceedings{liu2025picture,
  title={A picture is worth a thousand numbers: Enabling llms reason about time series via visualization},
  author={Liu, Haoxin and Liu, Chenghao and Prakash, B Aditya},
  booktitle={Proceedings of the 2025 Conference of the Nations of the Americas Chapter of the Association for Computational Linguistics: Human Language Technologies (Volume 1: Long Papers)},
  pages={7486--7518},
  year={2025}
}

@article{paszke2019pytorch,
  title={Pytorch: An imperative style, high-performance deep learning library},
  author={Paszke, Adam and Gross, Sam and Massa, Francisco and Lerer, Adam and Bradbury, James and Chanan, Gregory and Killeen, Trevor and Lin, Zeming and Gimelshein, Natalia and Antiga, Luca and others},
  journal={Advances in neural information processing systems},
  volume={32},
  year={2019}
}

@inproceedings{
loshchilov2018decoupled,
title={Decoupled Weight Decay Regularization},
author={Ilya Loshchilov and Frank Hutter},
booktitle={International Conference on Learning Representations},
year={2019},
}

@article{goyal2017accurate,
  title={Accurate, large minibatch sgd: Training imagenet in 1 hour},
  author={Goyal, Priya and Doll{\'a}r, Piotr and Girshick, Ross and Noordhuis, Pieter and Wesolowski, Lukasz and Kyrola, Aapo and Tulloch, Andrew and Jia, Yangqing and He, Kaiming},
  journal={arXiv preprint arXiv:1706.02677},
  year={2017}
}

\appendix
\section{Broader Impacts}

This work aims to improve the efficiency of LLM-based time-series analysis by reducing redundant TS and prompt tokens. 
The proposed method can lower inference latency and computational cost, which may benefit real-world applications that require efficient temporal modeling, such as healthcare monitoring, energy management, financial analysis, industrial maintenance, and climate-related forecasting. 
By reducing the token burden of LLM-based TS systems, TokenDecouple may also help decrease GPU memory consumption and energy usage, making such models more accessible in resource-constrained settings.

However, more efficient TS analysis systems may also amplify risks when deployed in high-stakes domains. 
For example, inaccurate predictions or reasoning results in healthcare, finance, or infrastructure monitoring could lead to harmful decisions if used without expert verification. 
In addition, compression-based methods may discard subtle but important temporal signals under severe distribution shifts or rare-event scenarios. 
Therefore, we recommend that TokenDecouple be used as an efficiency-enhancing component rather than an autonomous decision-making system. 
For safety-critical applications, model outputs should be accompanied by uncertainty estimation, domain-specific validation, and human oversight.

\section{Baselines}
\subsection{LLM-based TS Models}
\label{app:LLM-based TS Models}
\paragraph{OFA~\cite{zhou2023one}.}
OFA is an LLM-based time series model that transfers pretrained transformer backbones from NLP to time series analysis by freezing self-attention and feedforward layers while selectively fine-tuning embeddings and normalization layers. It redesigns the input embedding to project normalized and patched time series into the token space of pretrained models, enabling efficient cross-domain adaptation with minimal parameter updates. This design demonstrates that frozen pretrained transformers can be effectively repurposed for time series tasks through lightweight architectural modifications.

\paragraph{TimeLLM~\cite{jin2023time}.}
TimeLLM reprograms a frozen pretrained large language model, such as LLaMA or GPT-2, for time series forecasting without fine-tuning the backbone. It transforms normalized and patched time series into patch embeddings and aligns them with the language model via a patch reprogramming mechanism using a small set of learned text prototypes, enabling effective cross-modal alignment. Additional prompt-as-prefix conditioning is used to inject task instructions, dataset context, and input statistics, while only lightweight input transformation and output projection modules are trained.

\paragraph{CALF~\cite{liu2025calf}.}
CALF is an LLM-based time series model that adopts a dual-branch architecture consisting of a textual source branch and a temporal target branch with shared pretrained LLM weights. It aligns time series inputs with the language embedding space through a cross-modal match module based on principal word embeddings, while further enforcing alignment via intermediate feature regularization and output consistency losses. By explicitly matching representations across modalities, CALF aims to reduce modality gaps and better adapt pretrained language models to time series data.

\paragraph{S\textsuperscript{2}IP~\cite{pan2024s}.}
S\textsuperscript{2}IP is an LLM-based time series forecasting model that integrates time series tokenization with semantic space informed prompting. It decomposes normalized time series into trend, seasonal, and residual components, applies patching to obtain time series embeddings, and aligns them with a reduced set of semantic anchors derived from pretrained word embeddings. The top-$K$ relevant semantic anchors are retrieved as prefix prompts to guide a frozen LLM, while only lightweight input/output mappings and selected embedding components are fine-tuned.

\subsection{Token Merging Methods}
\label{app:Token Merging Methods}
\paragraph{Global Token Merging~\cite{casella2024statistical}.}
Global token merging for time series reduces the number of tokens at each layer by merging the most similar token pairs based on a global similarity metric, e.g., cosine similarity $s_{ij} = \frac{\mathbf{t}_i \cdot \mathbf{t}_j}{\|\mathbf{t}_i\|\|\mathbf{t}_j\|}$. At each merging step, token pairs with the highest similarity scores are combined, regardless of their temporal positions.

\paragraph{Local Token Merging~\cite{bolya2022token}.}
Local token merging for time series introduces locality constraints by restricting similarity computation and merging operations to adjacent or local temporal windows. Tokens are only merged if $|i-j| \leq w$, where $w$ denotes the window size. This design preserves temporal ordering and local dependency structures that are critical for time series modeling.

\paragraph{Dynamic Token Merging~\cite{feng2023efficient}.}
Dynamic token merging adaptively adjusts the merging ratio according to token similarity and model depth. Instead of using a fixed reduction rate, the number of merged tokens is determined by the distribution of similarity scores, allowing the effective token count $N_l$ at layer $l$ to vary with the input and representation complexity.

\section{Datasets}
\label{app:Datasets}
\subsection{Datasets for Imputation Task}
\paragraph{ETTh1~\cite{zhou2021informer}.} ETTh1 is a used public benchmark for long-term time series task, designed to model temperature variations and load dynamics in electric transformers. It contains 2 years of hourly observations collected from 2016/06 to 2018/06 , where each record includes a timestamp, the target variable oil temperature, and six load-related covariates. Following the standard setting, the dataset is split into 12 months for training, 4 months for validation, and 4 months for testing. 

\paragraph{ETTh2~\cite{zhou2021informer}.} ETTh2 is an industrial time series benchmark from the ETT dataset family released by Tsinghua University. It contains hourly observations from 2016 to 2018, with 17,420 samples and seven variables, including six power-load features and oil temperature as the target. Compared with ETTh1, ETTh2 is collected from a transformer station in a different region, where load fluctuations and temperature regulation patterns exhibit distinct regional characteristics. 

\paragraph{ETTm1~\cite{zhou2021informer}.} ETTm1 is a minute-level variant of the ETT benchmark for time series tasks. It contains two years of transformer operation records from July 2016 to July 2018, collected at 15-minute intervals. Each sample includes a timestamp, six power-load covariates, and oil temperature as the target. The load-related variables are HUFL, HULL, MUFL, MULL, LUFL, and LULL, corresponding to high-, middle-, and low-usage load measurements under different operating states. Compared with the hourly ETT datasets, ETTm1 provides finer-grained temporal observations, making it suitable for evaluating models’ ability to capture short-term fluctuations.

\paragraph{ETTm2~\cite{zhou2021informer}.} ETTm2 is minute-level benchmark from the ETT dataset family, designed for fine-grained electric transformer temperature forecasting. It contains two years of records from July 2016 to July 2018, collected at 15-minute intervals. Each record consists of a timestamp, six power-load measurements, and oil temperature as the prediction target. The six covariates, including HUFL, HULL, MUFL, MULL, LUFL, and LULL, describe high-, middle-, and low-usage load behaviors under different operating states. 

\paragraph{Weather~\cite{wu2022timesnet}.} Weather is a used benchmark dataset, containing 21 meteorological variables such as air temperature, humidity, wind speed, precipitation, and atmospheric pressure. The dataset records weather observations at 10-minute intervals over one year, providing fine-grained temporal dynamics for environmental modeling. Due to its diverse weather indicators and high sampling frequency, Weather is widely used to evaluate forecasting models’ ability to capture short-term fluctuations, periodic patterns, and complex dependencies among meteorological variables.

\paragraph{Electricity~\cite{li2019enhancing}.} Electricity is a public time series benchmark that records electricity consumption and related power measurements across customers or households. Depending on the setting, it contains either hourly consumption records from 2012 to 2014 for 321 customers, or minute-level household power records from 2006 to 2010. The dataset includes temporal fields such as date and time, together with variables including global active power, global reactive power, voltage, global intensity, and sub-metering values for different appliance groups.

\subsection{Datasets for Classification Task}
\paragraph{JapaneseVowels~\cite{kudo1999multidimensional}.} Japanese Vowels is a multivariate time series classification dataset constructed from speech recordings of nine male speakers pronouncing the Japanese vowel sequence /ae/. It contains 640 time series instances, with 270 samples for training and 370 samples for testing. Each time step is represented by 12 LPC cepstral coefficients, and the sequence length varies from 7 to 29. The recordings were sampled at 10 kHz, with a frame length of 25.6 ms and a frame shift of 6.4 ms. This dataset is commonly used to evaluate models on speaker or speech-pattern classification from short multivariate temporal signals.

\paragraph{Handwriting~\cite{shokoohi2017generalizing}.} Handwriting is a multivariate time series classification dataset collected from smartwatch motion signals while subjects write the 26 letters of the alphabet. It contains 1,000 instances, including 150 training samples and 850 test samples. Each sequence consists of three dimensions corresponding to the three-axis accelerometer readings from the smartwatch. The original sequences were padded to a consistent format, making the dataset suitable for evaluating models on gesture and motion-pattern classification from wearable sensor data.

\paragraph{Heartbeat~\cite{bagnall2018uea}.} Heartbeat is a multivariate time series dataset derived from the PhysioNet/CinC Challenge 2016 heart sound recordings. The data were collected from multiple contributors worldwide in both clinical and non-clinical environments, covering healthy subjects and patients with confirmed cardiac diagnoses. Each recording is labeled as either normal or abnormal, where abnormal cases are mainly associated with cardiac diseases such as valve defects and coronary artery disease. The recordings were truncated to 5 seconds and converted into spectrograms, where each dimension corresponds to a frequency band. The dataset contains two classes, with 113 normal instances and 296 abnormal instances, making it suitable for biomedical signal classification.

\paragraph{EthanolConcentration~\cite{large2018detecting}.} Ethanol Concentration is a multivariate time series classification dataset based on raw spectra of water–ethanol solutions measured from 44 real whisky bottles. The samples correspond to four ethanol concentration levels: 35\%, 38\%, 40\%, and 45\%. Since Scotch whisky is legally required to contain at least 40\% alcohol, the task is to classify the ethanol concentration of an unknown sample from its spectral signal. This dataset is commonly used to evaluate models on fine-grained classification of chemical or spectral time series data.

\paragraph{PEMS-SF~\cite{cuturi2011fast}.} PEMS-SF is a multivariate time series classification dataset derived from traffic sensor data provided by the California Department of Transportation. It contains 15 months of daily freeway lane-occupancy measurements from the San Francisco Bay Area, covering January 1, 2008 to March 30, 2009, sampled every 10 minutes. Each instance represents one day, with 963 sensor dimensions and a sequence length of 144. After removing public holidays and anomalous days, the dataset contains 440 daily time series. The classification task is to identify the corresponding day of the week, from Monday to Sunday, based on the observed traffic occupancy patterns.

\subsection{Datasets for Anomaly Detection Task}
\paragraph{SMD~\cite{su2019robust}.} SMD (Server Machine Dataset) is a public multivariate time series anomaly detection dataset released by the NetMan Lab at Tsinghua University for intelligent system monitoring. It contains five weeks of operation records from 28 server machines, where each machine is represented by 38 monitored variables. The data are collected at one-minute intervals, while the exact timestamp information is anonymized. Due to its machine-level monitoring signals and labeled abnormal segments, SMD is commonly used to evaluate anomaly detection methods under realistic server operation scenarios.

\paragraph{MSL~\cite{hundman2018detecting}.} MSL (Mars Science Laboratory) is a real-world time series anomaly detection dataset provided by NASA, derived from telemetry signals of the Curiosity rover. It consists of 27 sensor channels, each represented with 55 feature dimensions. The dataset is split into training and testing sets, where the training data contain only normal patterns and the test data include expert-labeled anomalies. Due to its spacecraft telemetry signals and realistic abnormal events, MSL is commonly used to evaluate anomaly detection methods under complex system monitoring scenarios.

\paragraph{SMAP~\cite{hundman2018detecting}.} SMAP (Soil Moisture Active Passive) is a real-world time series anomaly detection dataset provided by NASA, derived from spacecraft telemetry signals. It contains 55 telemetry channels with a total of 429,735 data points, including 69 expert-labeled anomalies. Similar to MSL, SMAP provides realistic system-monitoring signals with annotated abnormal events, making it commonly used for evaluating anomaly detection methods on multivariate spacecraft telemetry data.

\paragraph{SWaT~\cite{mathur2016swat}.} SWaT (Secure Water Treatment) is a real-world industrial control system dataset collected from a scaled water treatment testbed. It records the operation of a water treatment process under both normal conditions and cyber-attack scenarios. The dataset contains multivariate sensor and actuator signals from different stages of the treatment pipeline, reflecting physical process dynamics and control-system behaviors. Due to its realistic industrial setting and attack annotations, SWaT is commonly used for anomaly detection and security analysis in industrial control systems.

\paragraph{PSM~\cite{abdulaal2021practical}.} PSM (Pooled Server Metric Dataset) is a public time series anomaly detection dataset collected by eBay from multiple application server nodes. It contains 13 weeks of training data and 8 weeks of test data, with 132,481 records in total. Each record includes a timestamp and 24 server-performance metrics, such as CPU usage, memory usage, and other system-level indicators. With real server monitoring signals and anomaly annotations, PSM is commonly used for evaluating anomaly detection methods in large-scale service operation scenarios.

\paragraph{Traffic~\cite{wu2022timesnet}.} Traffic is a public time series benchmark that records hourly road occupancy rates from freeway sensors in the San Francisco Bay Area between 2015 and 2016. The dataset captures traffic dynamics across multiple sensors, reflecting temporal variations in road usage, congestion patterns, and daily or weekly mobility behaviors. It is commonly used for analyzing large-scale transportation time series and evaluating models on traffic-related temporal patterns.

\paragraph{Solar~\cite{lai2018modeling}.} Solar is a public renewable-energy time series benchmark containing solar power production records from 137 photovoltaic plants in Alabama State. The data were collected in 2006 at 10-minute intervals, capturing fine-grained variations in solar generation across different PV plants. Due to its strong daily periodicity, weather-related fluctuations, and multi-site generation patterns, the dataset is commonly used for analyzing temporal dynamics in solar energy production.

\paragraph{PEMS04~\cite{liu2022scinet}.} PEMS is a collection of public traffic-network time series datasets, including PEMS03, PEMS04, PEMS07, and PEMS08. These datasets record traffic measurements from road sensor networks and contain complex spatial-temporal dependencies across different locations. Due to their large-scale sensor structure and dynamic traffic patterns, PEMS datasets are widely used for analyzing urban traffic flow, road-network dynamics, and spatial-temporal modeling.

\paragraph{PEMS08~\cite{liu2022scinet}.} PEMS is a collection of public traffic-network time series datasets, including PEMS03, PEMS04, PEMS07, and PEMS08. These datasets record traffic measurements from road sensor networks and contain complex spatial-temporal dependencies across different locations. Due to their large-scale sensor structure and dynamic traffic patterns, PEMS datasets are widely used for analyzing urban traffic flow, road-network dynamics, and spatial-temporal modeling.

\paragraph{Exchange~\cite{wu2022timesnet}.} Exchange-Rate is a public financial time series benchmark containing daily exchange rates from 1990 to 2016. It covers eight economies, including Australia, Britain, Canada, Switzerland, China, Japan, New Zealand, and Singapore. The dataset records long-term currency dynamics and cross-country exchange-rate fluctuations, making it suitable for analyzing financial temporal patterns, trend changes, and interactions among multiple foreign exchange series.

\subsection{Datasets for Reasoning Task}
\paragraph{RCW~\cite{liu2025picture}.} RCW (North Atlantic Right Whale Calls) is an audio-based time series classification dataset used for detecting right whale calls in oceanic environmental noise. Each sample corresponds to an acoustic signal segment, and the task is to determine whether it contains a North Atlantic right whale call. The key discriminative cue is a short rising “whoop” sound, which provides clear signal-level evidence for the positive class. 

\paragraph{TEE~\cite{liu2025picture}.} TEE (Transient Electromagnetic Event) is a time series classification dataset derived from FORTE satellite observations. The satellite uses optical and radio-frequency instruments to detect transient electromagnetic events associated with lightning activity. Each class corresponds to a specific physical event type with distinguishable temporal characteristics in the measured signal. For example, the “CG Positive” category is related to positive cloud-to-ground discharge and is typically reflected as a sharp radiation spike followed by subsequent signal variations in the power-density time series.

\paragraph{ECG~\cite{liu2025picture}.} ECG is a single-lead electrocardiogram classification dataset covering four cardiac rhythm categories: normal sinus rhythm, atrial fibrillation, other alternative rhythms, and noisy or unclassifiable recordings. Unlike simple pattern-based tasks, each category depends on multiple clinical signal features. For example, atrial fibrillation is typically identified by combining cues such as irregular rhythm, missing P waves, abnormal baseline patterns, QRS characteristics, and fibrillatory waves.

\paragraph{EMG~\cite{liu2025picture}.} EMG is a single-channel electromyography classification dataset based on muscle responses to neural stimulation. It includes three categories: healthy, neuropathy, and myopathy. The task requires integrating multiple signal characteristics rather than relying on a single cue. For example, neuropathy is typically associated with fibrillations, positive sharp waves, high-amplitude and long-duration motor unit action potentials, and polyphasic waveforms.

\paragraph{HAR~\cite{liu2025picture}.} HAR is a human activity recognition dataset collected from smartphone wearable sensors. It records six daily activities, including walking, walking upstairs, walking downstairs, sitting, standing, and lying down. We use three sensor channels corresponding to body linear acceleration. Since the data were collected from 30 users with different motion habits, the same activity may exhibit user-specific variations, making HAR suitable for evaluating probabilistic reasoning over behavioral time series.

\paragraph{CTU~\cite{liu2025picture}.} CTU is a computer type usage dataset based on 24-hour electricity consumption patterns. The task is to infer whether the device is a desktop or a laptop from consumer electricity usage behavior. The dataset was collected from 251 users, whose individual usage habits introduce unobserved variability into the signals. 

\section{Dataset Statistics}
\paragraph{Imputation.}
For the imputation task, we evaluate the models on six widely used time series benchmarks, including ETTh1, ETTh2, ETTm1, ETTm2, Weather, and Electricity, whose statistics are reported in Tab.~\ref{tab:Imputation}. Following the standard protocol of TimesNet, we construct incomplete inputs by randomly masking $\{12.5\%, 25\%, 50\%\}$ of time points from time series segments with length 96. The models are required to reconstruct the missing values based on the remaining observed context. We compare each original model with its TokenDecouple-enhanced variant under all missing ratios, and report MSE and MAE as metrics.
\begin{table}[h]
\centering
\caption{Statistics of datasets used for imputation.}
\label{tab:Imputation}
\resizebox{\linewidth}{!}{
\begin{tabular}{l|ccc}
\toprule
\textbf{Dataset} & \textbf{Length} & \textbf{Dimension} & \textbf{Frequency} \\
\midrule
ETTh1        & 17420 & 7   & 1 hour \\
ETTh2        & 17420 & 7   & 1 hour \\
ETTm1       & 69680 & 7   & 15 min \\
ETTm2        & 69680 & 7   & 15 min \\
Weather     & 52696 & 22  & 10 min \\
Electricity & 26304 & 321 & 1 hour \\
\bottomrule
\end{tabular}
}
\end{table}

\paragraph{Classification.}
For the classification task, we evaluate the models on five multivariate time series classification benchmarks, including JapaneseVowels, Handwriting, Heartbeat, EthanolConcentration, and PEMS-SF. Their statistics are summarized in Tab.~\ref{tab:Classification}. These datasets cover diverse domains, including speech signals, wearable motion signals, biomedical recordings, chemical spectra, and traffic sensor measurements. We use Accuracy as the evaluation metric and compare each baseline model with its TokenDecouple-enhanced counterpart.

\begin{table}[h]
\centering
\caption{Statistics of datasets used for classification.}
\label{tab:Classification}
\resizebox{\linewidth}{!}{
\begin{tabular}{l|ccccc}
\toprule
\textbf{Dataset} & \textbf{Train Cases} & \textbf{Test Cases} & \textbf{Dimensions} & \textbf{Length} & \textbf{Classes} \\
\midrule
JapaneseVowels        & 270 & 370 & 12  & 29   & 9  \\
Handwriting           & 150 & 850 & 3   & 152  & 26 \\
Heartbeat             & 204 & 205 & 61  & 405  & 2  \\
EthanolConcentration  & 261 & 263 & 3   & 1751 & 4  \\
PEMS-SF               & 267 & 173 & 963 & 144  & 7  \\
\bottomrule
\end{tabular}
}
\end{table}

\begin{figure*}[t]
  \includegraphics[width=\textwidth]{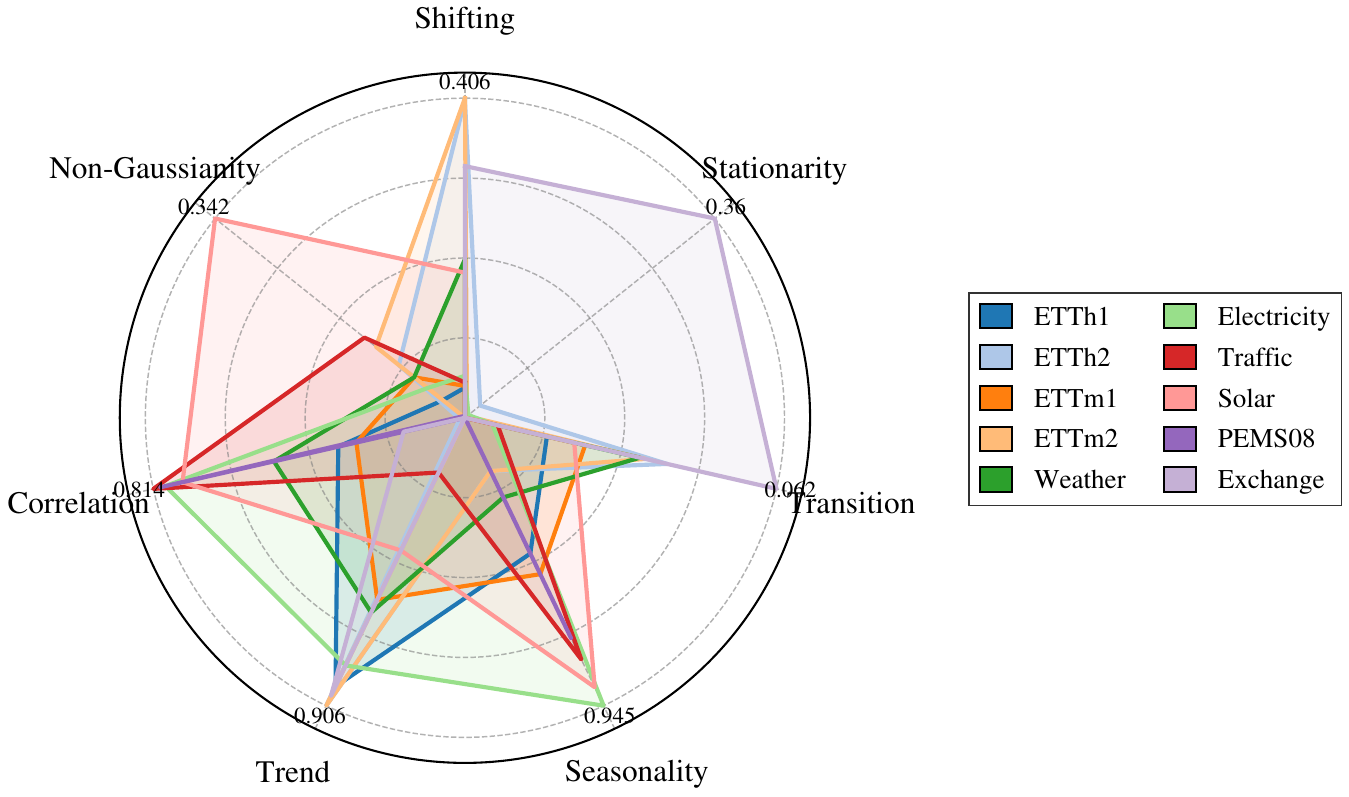}
    \caption{Statistical characteristics of datasets, including seasonality, trend, stationarity, transition, shifting, correlation, and non-Gaussianity.}
  \label{fig:dataset_characteristics_radar}
\end{figure*}

\paragraph{Anomaly Detection.}
For the anomaly detection task, we evaluate the models on three widely used multivariate time series anomaly detection benchmarks, including SMAP, MSL, and SMD. Their statistics are summarized in Tab.~\ref{tab:anomaly}. These datasets cover spacecraft telemetry and server-machine monitoring scenarios, where anomalies are identified from multivariate temporal signals. We use F1-score as the evaluation metric and compare each baseline model with its TokenDecouple-enhanced counterpart on all datasets.
\begin{table}[h]
\centering
\caption{Statistics of datasets used for anomaly detection.}
\label{tab:anomaly}
\resizebox{\linewidth}{!}{
\begin{tabular}{l|ccccc}
\toprule
\textbf{Dataset} & \textbf{Entities} & \textbf{Dimensions} & \textbf{Training Size} & \textbf{Testing Size} & \textbf{Anomaly Ratio (\%)} \\
\midrule
SMAP & 55 & 25 & 135,183 & 427,617 & 13.13 \\
MSL  & 27 & 55 & 58,317  & 73,729  & 10.72 \\
SMD  & 28 & 38 & 708,405 & 708,420 & 4.16  \\
\bottomrule
\end{tabular}
}
\end{table}

\paragraph{Forecasting.}
For the forecasting task, we evaluate the models on five widely used multivariate time series benchmarks, including Traffic, Solar, PEMS04, PEMS08, and Exchange. Their statistics are summarized in Tab.~\ref{tab:forecasting}. In the table, \textit{Dim} denotes the number of variables, \textit{Size} denotes the number of time points in the train, validation, and test splits, respectively, and \textit{Frequency} denotes the sampling interval. We adopt two prediction horizons $H \in \{192, 720\}$ and report MSE and MAE as evaluation metrics.
\begin{table}[h]
\centering
\caption{Statistics of datasets used for forecasting.}
\label{tab:forecasting}
\resizebox{\linewidth}{!}{
\begin{tabular}{l|ccc}
\toprule
\textbf{Dataset} & \textbf{Dim} & \textbf{Size (Train, Val, Test)} & \textbf{Frequency} \\
\midrule
Traffic  & 862 & (12185, 1757, 3509)  & Hourly \\
Solar    & 137 & (36601, 5161, 10417)  & 10 min \\
PEMS04   & 307 & (10172, 3375, 3375)   & 5 min \\
PEMS08   & 170 & (10690, 3548, 3548)   & 5 min \\
Exchange & 8   & (5120, 665, 1422)     & Daily \\
\bottomrule
\end{tabular}
}
\end{table}

\paragraph{Reasoning-centric Tasks.}
We evaluate reasoning tasks across three levels of complexity, including Simple Deterministic Reasoning, Complex Deterministic Reasoning, and Probabilistic Reasoning. The dataset statistics are summarized in Table~\ref{tab:reasoning_dataset_statistics}. Simple Deterministic Reasoning includes RCW and TEE, where class labels correspond to clearly defined temporal patterns annotated by human experts. Complex Deterministic Reasoning includes ECG record diagnosis and EMG signal diagnosis, where labels require the joint interpretation of multiple temporal patterns by medical experts. Probabilistic Reasoning includes HAR and CTU, where class boundaries are less deterministic due to uncertain feature--label relationships and unobserved user-specific behaviors.

\begin{table}[h]
\centering
\caption{Statistics of datasets used for reasoning tasks.}
\label{tab:reasoning_dataset_statistics}
\resizebox{\linewidth}{!}{
\begin{tabular}{l|cccc}
\toprule
\textbf{Dataset} & \textbf{Variables} & \textbf{Length} & \textbf{Classes} & \textbf{Samples} \\
\midrule
RCW & 1 & 4000 & 2 & 30,000 \\
TEE & 1 & 319  & 7 & 143 \\
ECG & 1 & 1500 & 4 & 43,673 \\
EMG & 1 & 1500 & 3 & 205 \\
CTU & 1 & 720  & 2 & 500 \\
HAR & 3 & 206  & 6 & 10,299 \\
\bottomrule
\end{tabular}
}
\end{table}

\paragraph{Statistical Characteristics of Datasets.}
We characterize each dataset from seven statistical perspectives, including \textbf{shifting}, \textbf{stationarity}, \textbf{transition}, \textbf{seasonality}, \textbf{trend}, \textbf{correlation}, and \textbf{non-Gaussianity}. 
Shifting measures the degree of distributional variation across time, while stationarity reflects whether the statistical properties of the sequence remain stable. 
Transition captures abrupt changes or regime switches in temporal dynamics. 
Seasonality and trend describe periodic structures and long-term directional movements, respectively. 
Correlation reflects the dependency among variables, and non-Gaussianity measures the deviation of the data distribution from a Gaussian assumption. 
Together, these attributes provide a comprehensive description of the temporal complexity and distributional diversity of different datasets.

As shown in Fig.~\ref{fig:dataset_characteristics_radar}, the datasets exhibit diverse statistical properties. 
For example, ETT datasets show relatively strong trend patterns, Electricity and Solar contain clear seasonality, Traffic and PEMS08 exhibit strong inter-variable correlations, while Exchange presents more evident stationarity variation and transition behavior. 
This diversity indicates that the evaluation covers a broad range of temporal regimes, rather than being restricted to datasets with similar dynamics. 
Therefore, the experimental results reflect model behavior under heterogeneous temporal structures and distributional conditions.

The broad coverage of dataset characteristics also helps assess the robustness of our method. 
Despite substantial differences in shifting, seasonality, trend, correlation, and distributional shape, our method maintains stable performance gains across datasets. 
This suggests that the proposed token decoupling strategy is not overly dependent on a specific statistical pattern or dataset type. 
Instead, by separately compressing TS tokens and prompt tokens according to their distinct functional roles, the method remains effective under diverse temporal dynamics and is less sensitive to dataset-specific statistical variations.

\section{Theoretical Analysis of TS Token Merging}
\label{app:ts_token_merging_theory}

This section provides a stability analysis for the proposed frequency-domain TS token merging strategy. 
The goal is not to prove that token merging is lossless or always improves task performance. 
Instead, we show that the perturbation introduced by TS token merging is explicitly bounded under mild assumptions. 
Specifically, we prove that: (i) frequency-domain perturbation is equivalent to time-domain token perturbation under an orthonormal DFT; (ii) active spectral token selection controls the residual energy of inactive tokens; (iii) local affinity-based merging controls the group-wise merging error; and (iv) under the Lipschitz stability of the LLM backbone and task head, the induced output perturbation is also bounded. 
This analysis explains why frequency-domain TS token merging can reduce token redundancy while preserving dominant temporal information.

\paragraph{Setup.}
Let $X_b\in\mathbb{R}^{N\times D}$ denote the TS token sequence of the $b$-th sample, where $N$ is the number of TS tokens and $D$ is the token dimension. 
We apply an orthonormal DFT along the token dimension and obtain $F_b=\mathcal{F}X_b\in\mathbb{C}^{N\times D}$, where $\mathcal{F}$ is a unitary DFT matrix satisfying $\mathcal{F}^{*}\mathcal{F}=I$. 
For frequency component $k$, let $F_{b,i,k}\in\mathbb{C}^{D}$ denote the spectral vector of token $i$ and define its spectral energy as $\|F_{b,i,k}\|_2^2$.

For each frequency $k$, the active token set $\mathcal{I}_{b,k}$ is selected to preserve at least a $\rho$ fraction of the spectral energy, i.e., $\sum_{i\in\mathcal{I}_{b,k}}\|F_{b,i,k}\|_2^2 \geq \rho\sum_{i=1}^{N}\|F_{b,i,k}\|_2^2$. 
The active set is partitioned into local merging groups $\{\mathcal{G}_{b,k}^{(u)}\}_{u=1}^{m_{b,k}}$. 
For each group, the merged representative is defined as $\tilde F_{b,u,k}=\sum_{i\in\mathcal{G}_{b,k}^{(u)}}\omega_{b,i}^{(u,k)}F_{b,i,k}$, where $\omega_{b,i}^{(u,k)}\geq 0$ and $\sum_{i\in\mathcal{G}_{b,k}^{(u)}}\omega_{b,i}^{(u,k)}=1$. 
Thus, the merged token is a convex combination of tokens within the same local spectral group.

Since the compressed representation has fewer tokens than the original one, we introduce an expanded merged representation only for theoretical comparison. 
For every $i\in\mathcal{G}_{b,k}^{(u)}$, we set $\bar F_{b,i,k}=\tilde F_{b,u,k}$. 
The resulting $\bar F_b$ has the same index structure as $F_b$, and $\bar X_b=\mathcal{F}^{-1}\bar F_b$ denotes its inverse-DFT counterpart. 
This expanded representation is only used in the proof and does not change the actual compressed inference process.

\paragraph{Lemma 1 (Frequency-domain perturbation equals time-domain perturbation).}
If $\mathcal{F}$ is an orthonormal DFT matrix, then the token-level perturbation in the time domain equals the perturbation in the frequency domain:
\[
\|X_b-\bar X_b\|_F=\|F_b-\bar F_b\|_F .
\]

\textit{
Proof. Since $F_b=\mathcal{F}X_b$ and $\bar F_b=\mathcal{F}\bar X_b$, we have $F_b-\bar F_b=\mathcal{F}(X_b-\bar X_b)$. 
Because $\mathcal{F}^{*}\mathcal{F}=I$, the Frobenius norm is preserved under the orthonormal DFT. 
Therefore, $\|F_b-\bar F_b\|_F=\|X_b-\bar X_b\|_F$. 
This shows that controlling the merging error in the frequency domain is equivalent to controlling the perturbation in the original TS token space.
}

\paragraph{Lemma 2 (Inactive spectral residual is bounded).}
For each frequency component $k$, if the active token set $\mathcal{I}_{b,k}$ preserves at least a $\rho$ fraction of the spectral energy, then the residual energy of inactive tokens is bounded by
\[
\sum_{k=0}^{K-1}\sum_{i\notin\mathcal{I}_{b,k}}\|F_{b,i,k}\|_2^2
\leq
(1-\rho)\|F_b\|_F^2 .
\]

\textit{
Proof. By the definition of $\mathcal{I}_{b,k}$, we have $\sum_{i\in\mathcal{I}_{b,k}}\|F_{b,i,k}\|_2^2 \geq \rho\sum_{i=1}^{N}\|F_{b,i,k}\|_2^2$. 
Hence, the inactive part at frequency $k$ satisfies $\sum_{i\notin\mathcal{I}_{b,k}}\|F_{b,i,k}\|_2^2 \leq (1-\rho)\sum_{i=1}^{N}\|F_{b,i,k}\|_2^2$. 
Summing this inequality over all frequencies gives the stated bound. 
This result shows that active token selection directly controls the spectral information discarded by merging.
}

\paragraph{Lemma 3 (Local spectral merging error is bounded).}
For each local group $\mathcal{G}_{b,k}^{(u)}$, define its spectral diameter as $\delta_{b,k}^{(u)}=\max_{i,j\in\mathcal{G}_{b,k}^{(u)}}\|F_{b,i,k}-F_{b,j,k}\|_2$. 
Then replacing all tokens in this group by their merged representative introduces the following bounded error:
\[
\sum_{i\in\mathcal{G}_{b,k}^{(u)}}\|F_{b,i,k}-\bar F_{b,i,k}\|_2^2
\leq
|\mathcal{G}_{b,k}^{(u)}|(\delta_{b,k}^{(u)})^2 .
\]

\textit{
Proof. For any $i\in\mathcal{G}_{b,k}^{(u)}$, since $\bar F_{b,i,k}=\tilde F_{b,u,k}$ and $\tilde F_{b,u,k}$ is a convex combination of tokens in the same group, we have
}
\[
\begin{aligned}
\|F_{b,i,k}-\bar F_{b,i,k}\|_2
=
\left\|
F_{b,i,k}
-
\sum_{j\in\mathcal{G}_{b,k}^{(u)}}
\omega_{b,j}^{(u,k)}F_{b,j,k}
\right\|_2 \\
\leq
\sum_{j\in\mathcal{G}_{b,k}^{(u)}}
\omega_{b,j}^{(u,k)}
\|F_{b,i,k}-F_{b,j,k}\|_2
\leq
\delta_{b,k}^{(u)} .
\end{aligned}
\]
\textit{
Squaring the inequality and summing over all tokens in $\mathcal{G}_{b,k}^{(u)}$ gives the result. 
Therefore, the local merging error is controlled by the spectral diameter of the group, which justifies restricting merging to locally similar TS tokens.
}

\paragraph{Theorem 1 (Frequency-domain TS token merging induces bounded output perturbation).}
Assume that the LLM backbone $f_{\Theta}$ is $L_f$-Lipschitz continuous and the task-specific head $h$ is $L_h$-Lipschitz continuous under the Frobenius norm. 
Then the final output perturbation induced by frequency-domain token merging is bounded by:
\[
\begin{aligned}
\|h(f_{\Theta}(X_b))-h(f_{\Theta}(\bar X_b))\|_F 
\leq \\
L_hL_f
\left[
\sum_{k=0}^{K-1}
\sum_{u=1}^{m_{b,k}}
|\mathcal{G}_{b,k}^{(u)}|
(\delta_{b,k}^{(u)})^2
+
(1-\rho)\|F_b\|_F^2
\right]&^{1/2}.
\end{aligned}
\]

\textit{
Proof. We first decompose the frequency-domain perturbation into the active merging error and the inactive residual. 
By Lemma 3, the active merging error over all groups and frequencies is bounded by $\sum_{k=0}^{K-1}\sum_{u=1}^{m_{b,k}}|\mathcal{G}_{b,k}^{(u)}|(\delta_{b,k}^{(u)})^2$. 
By Lemma 2, the inactive spectral residual is bounded by $(1-\rho)\|F_b\|_F^2$. 
Therefore,
}
\[
\begin{aligned}
\|F_b-\bar F_b\|_F^2
\leq\\
\sum_{k=0}^{K-1}
\sum_{u=1}^{m_{b,k}}
|\mathcal{G}_{b,k}^{(u)}|
&(\delta_{b,k}^{(u)})^2
+
(1-\rho)\|F_b\|_F^2 .
\end{aligned}
\]
\textit{
Using Lemma 1, we further have $\|X_b-\bar X_b\|_F=\|F_b-\bar F_b\|_F$. 
Since $f_{\Theta}$ and $h$ are Lipschitz continuous, the output perturbation satisfies:
}
\[
\begin{aligned}
\|h(f_{\Theta}(X_b))-h(f_{\Theta}(\bar X_b))\|_F\\
\leq
L_h\|f_{\Theta}(X_b)-f_{\Theta}(\bar X_b)\|_F \\
\leq
L_hL_f\|X_b-\bar X_b\|_F \\
=
L_hL_f\|F_b-\bar F_b\|_F .
\end{aligned}
\]
\textit{
Substituting the frequency-domain perturbation bound into the above inequality gives the theorem.
}

\paragraph{Discussion.}
Theorem 1 shows that the effect of TS token merging is governed by two interpretable terms. 
The first term, $\sum_{k,u}|\mathcal{G}_{b,k}^{(u)}|(\delta_{b,k}^{(u)})^2$, is the local merging error. 
It becomes small when tokens within each group have similar spectral structures, which supports the use of the local affinity graph. 
The second term, $(1-\rho)\|F_b\|_F^2$, is the inactive spectral residual. 
It is controlled by the retained spectral energy ratio $\rho$, which explains the role of active spectral token selection. 
Therefore, the proposed TS token merging strategy is not arbitrary compression: it preserves dominant spectral components and merges only locally similar tokens, leading to a bounded perturbation on the output.

This theoretical result directly supports the design of the TS token merging module in TokenDecouple. 
The DFT step provides a space where spectral redundancy can be measured; active token selection preserves dominant frequency information; local affinity-based grouping controls the merging error; and the adaptive merging budget adjusts the compression strength across frequencies. 
Together, these components reduce the number of TS tokens while keeping the induced representation and output perturbations bounded.
\section{Computational Cost of Prompt Tokens}
We consider an $L$-layer LLM with an initial number of prompt tokens denoted as $N_{\text{Text}}$. To reduce the computational overhead introduced by prompt tokens, we adopt a pyramidal decay strategy across layers, where the number of prompt tokens decreases geometrically with model depth.

Specifically, we assume that the prompt token count is reduced by a constant ratio of $0.25$ at each successive layer. As a result, the number of prompt tokens processed at layer $l$ is given by:
\[
N_{\text{Text}}^{(l)} = 0.25^{\,l} N_{\text{Text}},
\]
which produces a rapidly decaying sequence of approximately $100\%$, $25\%$, $6.3\%$, $1.6\%$, and $0.4\%$ of the original prompt length across early layers.

The total number of prompt tokens processed throughout the entire network is obtained by summing the contributions from all layers:
\[
\sum_{l=1}^{L} N_{\text{Text}}^{(l)} = \sum_{l=1}^{L} 0.25^{\,l} N_{\text{Text}}.
\]
This summation forms a finite geometric series and can be evaluated in closed form as:
\[
\sum_{l=1}^{L} N_{\text{Text}}^{(l)} = N_{\text{Text}} \frac{1 - 0.25^{L}}{1 - 0.25}.
\]

To characterize the effective prompt length experienced by each layer on average, we divide the total prompt token count by the number of layers $L$, yielding:
\[
\bar{N}_{\text{Text}} = \frac{1}{L} \sum_{l=1}^{L} N_{\text{Text}}^{(l)}
= \frac{N_{\text{Text}}}{L(1 - 0.25)} (1 - 0.25^{L}).
\]

For moderately deep language models (i.e., $L \ge 4$), the term $0.25^{L}$ rapidly approaches zero and can be safely neglected. Under this approximation, the average number of prompt tokens per layer simplifies to:
\[
\bar{N}_{\text{Text}} \approx \frac{4}{3L} N_{\text{Text}}.
\]

We then define the prompt token compression ratio as:
\[
\eta_{L} = 1 - \frac{\bar{N}_{\text{Text}}}{N_{\text{Text}}},
\]
which leads to:
\[
\eta_{L} \approx 1 - \frac{4}{3L}.
\]

This analysis shows that prompt-induced computational overhead decreases sharply with model depth. For example, a 12-layer LLM achieves a compression ratio of approximately $\eta_{12} \approx 89\%$, while a 24-layer model reaches $\eta_{24} \approx 94\%$, demonstrating that deeper models naturally amortize prompt costs under pyramidal token decay.
\section{Layer-wise Prompt Contribution Analysis}
\label{app:prompt_contribution}

To understand the layer-wise behavior of prompt token compression, we analyze the actual contribution of prompt tokens to TS tokens. 
Motivated by this, we first examine the interaction between TS tokens and prompt tokens in Transformer attention. 
Let $A_l^h$ denote the attention matrix of the $h$-th attention head at layer $l$. 
Let $\mathcal{T}$ and $\mathcal{P}$ denote the index sets of TS tokens and prompt tokens, respectively. 
For any TS token $i\in\mathcal{T}$, its total attention assigned to all prompt tokens is defined as:
\[
C_{l,h}(i)=\sum_{j\in\mathcal{P}} A_l^h(i,j).
\]
This quantity measures the proportion of information absorbed by TS token $i$ from prompt tokens when updating its representation in the $h$-th head of layer $l$. 
A larger $C_{l,h}(i)$ indicates that prompt tokens still exert a strong influence on TS representation learning at this layer, whereas a smaller value suggests that TS tokens rely less on prompt tokens.
To characterize the layer-wise trend, we define the Prompt Contribution Ratio as:
\[
\mathrm{PCR}_l
=
\frac{1}{H|\mathcal{T}|}
\sum_{h=1}^{H}
\sum_{i\in\mathcal{T}}
\sum_{j\in\mathcal{P}}
A_l^h(i,j),
\]
where $H$ is the number of attention heads and $|\mathcal{T}|$ is the number of TS tokens. 
$\mathrm{PCR}_l$ measures the average attention mass assigned by TS tokens to prompt tokens at layer $l$. 
A higher $\mathrm{PCR}_l$ indicates a stronger contribution of prompt tokens to TS representation updates, while a lower value implies a weaker marginal effect of prompt tokens.

However, $\mathrm{PCR}_l$ alone is insufficient to determine the compression ratio, since it only captures the contribution strength of prompt tokens but not the token reduction induced by the compression strategy. 
Therefore, we further introduce the Prompt Retention Ratio to describe how many prompt tokens are preserved at each layer under different compression ratios. 
Let $P_0$ denote the number of prompt tokens at the input layer and $P_l$ denote the number of prompt tokens retained at layer $l$. 
The retention ratio is defined as:
\[
R_l(\alpha)=\frac{P_l}{P_0}.
\]
When a fixed hierarchical compression ratio $\alpha$ is used, the number of prompt tokens is updated by
\[
P_{l+1}=\lceil \alpha P_l\rceil.
\]
Ignoring the rounding effect, we have
\[
R_l(\alpha)\approx \alpha^l.
\]

\begin{figure}[t]
  \includegraphics[width=\columnwidth]{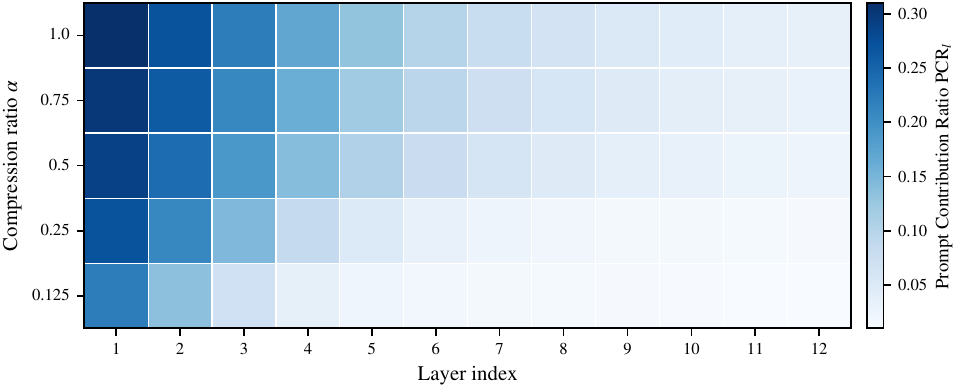}
  \caption{
  Layer-wise Prompt Contribution Ratio over the first 12 layers under different prompt compression ratios. 
  Higher values indicate stronger attention from TS tokens to prompt tokens, showing that prompt contribution is mainly concentrated in shallow layers and rapidly decays with depth.
  }
  \label{fig:prompt_pcr}
\end{figure}

We further analyze the layer-wise behavior of prompt token compression over the first 12 layers. 
Specifically, Fig.~\ref{fig:prompt_pcr} reports the Layer-wise Prompt Contribution Ratio, which measures the average attention mass assigned by TS tokens to prompt tokens at each layer. 
The results show that prompt contribution is relatively high in the shallow layers but decreases rapidly as the model depth increases. 
This indicates that prompt tokens mainly provide semantic guidance in the early stages, where TS tokens absorb task instructions, contextual information, and answer-related constraints. 
After several shallow layers, the direct contribution of prompt tokens becomes marginal, suggesting that keeping full prompt tokens in deeper layers introduces substantial redundant computation.

\begin{figure}[t]
  \includegraphics[width=\columnwidth]{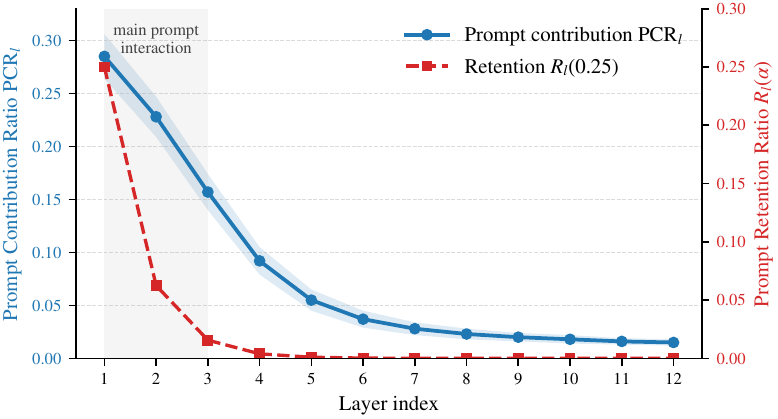}
  \caption{
  Layer-wise comparison between Prompt Contribution Ratio and Prompt Retention Ratio. 
  The contribution of prompt tokens decreases rapidly across layers, while $\alpha=0.25$ provides a matching retention decay pattern that preserves shallow-layer semantic guidance and reduces deep-layer prompt redundancy.
  }
  \label{fig:prompt_retention}
\end{figure}

To further relate this observation to the compression strategy, Fig.~\ref{fig:prompt_retention} shows the Prompt Retention Ratio under different hierarchical compression ratios. 
The retention curves indicate that larger ratios such as $\alpha=0.5$ and $\alpha=0.75$ preserve many prompt tokens even in deeper layers, which is inconsistent with the rapidly decaying prompt contribution observed in Fig.~\ref{fig:prompt_pcr}. 
In contrast, an overly small ratio such as $\alpha=0.125$ removes prompt tokens too aggressively and may reduce prompt information before it is sufficiently absorbed. 
The setting $\alpha=0.25$ provides a more balanced decay pattern: it preserves prompt tokens in the shallow layers while quickly reducing them after the main contribution window. 
These results support the design of hierarchical prompt compression, showing that prompt tokens should be progressively compressed according to their layer-wise contribution rather than fully retained across all layers.

\section{Implementation Details}
During training, we optimize all models using the AdamW optimizer~\cite{loshchilov2018decoupled} and adopt a cosine learning rate schedule for learning rate decay. To improve training stability, we apply a warmup strategy at the early training stage, with the warmup ratio set to $0.03$~\cite{goyal2017accurate}. Mixed precision training is used to reduce GPU memory consumption and improve training efficiency. For each group of comparative experiments, we keep the batch size, number of training epochs, and optimizer settings consistent across all compared methods. The main hyperparameters include learning rate, weight decay, batch size, maximum number of epochs, gradient clipping threshold, and dropout ratio. For TokenDecouple, the compression regularization term $\mathcal{L}_{\mathrm{pen}}$ is jointly optimized with the main task loss, where the coefficient $\lambda$ controls the trade-off between compression strength and task performance.

During inference, TokenDecouple enables both TS token merging and prompt token compression. TS token merging reduces the number of time-series tokens at the input stage, thereby lowering the computational cost of subsequent self-attention operations. Prompt token compression progressively reduces prompt tokens in the shallow layers of the LLM, preventing unnecessary attention computation over redundant tokens in deeper layers.

All experiments are conducted on the same hardware platform. We use 2 NVIDIA A800 GPUs for both training and inference, and implement all models based on PyTorch~\cite{paszke2019pytorch} and HuggingFace Transformers. For large-scale model training, we employ mixed precision and gradient accumulation to accommodate GPU memory constraints. All stochastic processes, including data splitting, parameter initialization, and training sampling, are controlled with fixed random seeds. To ensure the reliability of the results, the main experiments are repeated under multiple random seeds, and the average performance is reported.
\section{Full Results}
We provide the complete experimental results for all pattern-centric tasks in Tab.~\ref{tab:imputation_mae}--\ref{tab:forecasting_mse}. Specifically, Tab.~\ref{tab:imputation_mae} and~\ref{tab:imputation_mse} report the MAE and MSE results for the imputation task under three missing ratios, i.e., $12.5\%$, $25\%$, and $50\%$. Tab.~\ref{tab:classification_acc} presents the Accuracy results for the classification task, while Tab.~\ref{tab:anomaly_f1} reports the F1-score results for the anomaly detection task. For the forecasting task, Tab.~\ref{tab:forecasting_mae} and~\ref{tab:forecasting_mse} provide the MAE and MSE results under two prediction horizons, i.e., $H=192$ and $H=720$. Across these tables, we compare the original models (Ori) with their TokenDecouple-enhanced counterparts (TD), and report the mean performance with standard deviations over five random seeds.

\begin{table*}[t]
\centering
\caption{MAE results on the imputation task under 3 missing ratios(12.5\%, 25\%, and 50\%), comparing the original models (Ori) and TokenDecouple (TD) on pattern-centric tasks.  The results are obtained from 5 random seeds.}
\label{tab:imputation_mae}
\resizebox{\textwidth}{!}{
\begin{tabular}{ll|cc|cc|cc|cc}
\toprule
\multirow{2}{*}{\textbf{Dataset}} 
& \multirow{2}{*}{\textbf{Ratio}} 
& \multicolumn{2}{c|}{\textbf{OFA}} 
& \multicolumn{2}{c|}{\textbf{TimeLLM}} 
& \multicolumn{2}{c|}{\textbf{CALF}} 
& \multicolumn{2}{c}{\textbf{S\textsuperscript{2}IP}} \\
\cmidrule(lr){3-4} \cmidrule(lr){5-6} \cmidrule(lr){7-8} \cmidrule(lr){9-10}
& & \textbf{Ori} & \textbf{TD} 
  & \textbf{Ori} & \textbf{TD} 
  & \textbf{Ori} & \textbf{TD} 
  & \textbf{Ori} & \textbf{TD} \\
\midrule
\multirow{3}{*}{ETTh1}
& 12.5\% & $0.136{\pm}0.003$ & $0.134{\pm}0.001$ & $0.103{\pm}0.004$ & $0.099{\pm}0.002$ & $0.249{\pm}0.001$ & $0.244{\pm}0.003$ & $0.138{\pm}0.000$ & $0.133{\pm}0.002$ \\
& 25\%   & $0.151{\pm}0.002$ & $0.147{\pm}0.004$ & $0.095{\pm}0.001$ & $0.096{\pm}0.003$ & $0.301{\pm}0.004$ & $0.296{\pm}0.002$ & $0.131{\pm}0.003$ & $0.124{\pm}0.001$ \\
& 50\%   & $0.211{\pm}0.004$ & $0.205{\pm}0.002$ & $0.070{\pm}0.000$ & $0.065{\pm}0.001$ & $0.356{\pm}0.003$ & $0.350{\pm}0.004$ & $0.135{\pm}0.002$ & $0.122{\pm}0.003$ \\
\midrule
\multirow{3}{*}{ETTh2}
& 12.5\% & $0.117{\pm}0.001$ & $0.114{\pm}0.003$ & $0.137{\pm}0.002$ & $0.129{\pm}0.004$ & $0.324{\pm}0.003$ & $0.311{\pm}0.001$ & $0.116{\pm}0.000$ & $0.118{\pm}0.002$ \\
& 25\%   & $0.127{\pm}0.004$ & $0.123{\pm}0.001$ & $0.135{\pm}0.003$ & $0.133{\pm}0.002$ & $0.338{\pm}0.004$ & $0.319{\pm}0.003$ & $0.104{\pm}0.001$ & $0.102{\pm}0.004$ \\
& 50\%   & $0.150{\pm}0.002$ & $0.147{\pm}0.000$ & $0.152{\pm}0.004$ & $0.153{\pm}0.002$ & $0.472{\pm}0.001$ & $0.458{\pm}0.004$ & $0.091{\pm}0.003$ & $0.099{\pm}0.001$ \\
\midrule
\multirow{3}{*}{ETTm1}
& 12.5\% & $0.079{\pm}0.002$ & $0.077{\pm}0.001$ & $0.098{\pm}0.003$ & $0.097{\pm}0.000$ & $0.368{\pm}0.004$ & $0.376{\pm}0.002$ & $0.085{\pm}0.001$ & $0.088{\pm}0.003$ \\
& 25\%   & $0.090{\pm}0.004$ & $0.091{\pm}0.002$ & $0.062{\pm}0.001$ & $0.068{\pm}0.003$ & $0.294{\pm}0.002$ & $0.286{\pm}0.004$ & $0.089{\pm}0.000$ & $0.087{\pm}0.001$ \\
& 50\%   & $0.122{\pm}0.003$ & $0.118{\pm}0.004$ & $0.053{\pm}0.001$ & $0.054{\pm}0.002$ & $0.256{\pm}0.003$ & $0.239{\pm}0.001$ & $0.103{\pm}0.004$ & $0.098{\pm}0.002$ \\
\midrule
\multirow{3}{*}{ETTm2}
& 12.5\% & $0.073{\pm}0.001$ & $0.069{\pm}0.004$ & $0.075{\pm}0.002$ & $0.069{\pm}0.001$ & $0.304{\pm}0.003$ & $0.293{\pm}0.000$ & $0.041{\pm}0.002$ & $0.042{\pm}0.001$ \\
& 25\%   & $0.076{\pm}0.003$ & $0.070{\pm}0.001$ & $0.075{\pm}0.004$ & $0.071{\pm}0.002$ & $0.273{\pm}0.001$ & $0.266{\pm}0.003$ & $0.043{\pm}0.000$ & $0.046{\pm}0.004$ \\
& 50\%   & $0.089{\pm}0.004$ & $0.083{\pm}0.002$ & $0.058{\pm}0.000$ & $0.062{\pm}0.003$ & $0.278{\pm}0.002$ & $0.278{\pm}0.004$ & $0.045{\pm}0.001$ & $0.044{\pm}0.003$ \\
\midrule
\multirow{3}{*}{Weather}
& 12.5\% & $0.044{\pm}0.002$ & $0.040{\pm}0.001$ & $0.070{\pm}0.004$ & $0.065{\pm}0.003$ & $0.111{\pm}0.001$ & $0.116{\pm}0.004$ & $0.045{\pm}0.000$ & $0.041{\pm}0.002$ \\
& 25\%   & $0.047{\pm}0.003$ & $0.042{\pm}0.004$ & $0.031{\pm}0.001$ & $0.034{\pm}0.000$ & $0.115{\pm}0.003$ & $0.112{\pm}0.002$ & $0.046{\pm}0.001$ & $0.041{\pm}0.003$ \\
& 50\%   & $0.060{\pm}0.004$ & $0.058{\pm}0.002$ & $0.029{\pm}0.003$ & $0.022{\pm}0.001$ & $0.139{\pm}0.004$ & $0.133{\pm}0.003$ & $0.042{\pm}0.002$ & $0.037{\pm}0.000$ \\
\midrule
\multirow{3}{*}{ECL}
& 12.5\% & $0.185{\pm}0.001$ & $0.180{\pm}0.003$ & $0.079{\pm}0.002$ & $0.081{\pm}0.004$ & $0.221{\pm}0.003$ & $0.216{\pm}0.001$ & $0.171{\pm}0.004$ & $0.171{\pm}0.002$ \\
& 25\%   & $0.196{\pm}0.004$ & $0.197{\pm}0.002$ & $0.081{\pm}0.001$ & $0.080{\pm}0.003$ & $0.226{\pm}0.004$ & $0.214{\pm}0.000$ & $0.185{\pm}0.002$ & $0.177{\pm}0.004$ \\
& 50\%   & $0.211{\pm}0.003$ & $0.213{\pm}0.001$ & $0.076{\pm}0.004$ & $0.073{\pm}0.002$ & $0.265{\pm}0.001$ & $0.246{\pm}0.003$ & $0.206{\pm}0.004$ & $0.186{\pm}0.001$ \\
\bottomrule
\end{tabular}
}
\end{table*}

\begin{table*}[t]
\centering
\caption{MSE results on the imputation task under 3 missing ratios (12.5\%, 25\%, and 50\%), comparing the original models (Ori) and TokenDecouple (TD) on pattern-centric tasks. The results are obtained from 5 random seeds.}
\label{tab:imputation_mse}
\resizebox{\textwidth}{!}{
\begin{tabular}{ll|cc|cc|cc|cc}
\toprule
\multirow{2}{*}{\textbf{Dataset}} 
& \multirow{2}{*}{\textbf{Ratio}} 
& \multicolumn{2}{c|}{\textbf{OFA}} 
& \multicolumn{2}{c|}{\textbf{TimeLLM}} 
& \multicolumn{2}{c|}{\textbf{CALF}} 
& \multicolumn{2}{c}{\textbf{S\textsuperscript{2}IP}} \\
\cmidrule(lr){3-4} \cmidrule(lr){5-6} \cmidrule(lr){7-8} \cmidrule(lr){9-10}
& & \textbf{Ori} & \textbf{TD} 
  & \textbf{Ori} & \textbf{TD} 
  & \textbf{Ori} & \textbf{TD} 
  & \textbf{Ori} & \textbf{TD} \\
\midrule
\multirow{3}{*}{ETTh1}
& 12.5\% & $0.039{\pm}0.002$ & $0.037{\pm}0.001$ & $0.024{\pm}0.003$ & $0.024{\pm}0.000$ & $0.265{\pm}0.002$ & $0.262{\pm}0.003$ & $0.042{\pm}0.001$ & $0.035{\pm}0.002$ \\
& 25\%   & $0.049{\pm}0.003$ & $0.048{\pm}0.001$ & $0.016{\pm}0.001$ & $0.015{\pm}0.002$ & $0.274{\pm}0.003$ & $0.270{\pm}0.001$ & $0.039{\pm}0.002$ & $0.037{\pm}0.003$ \\
& 50\%   & $0.099{\pm}0.001$ & $0.097{\pm}0.003$ & $0.008{\pm}0.000$ & $0.008{\pm}0.001$ & $0.370{\pm}0.002$ & $0.354{\pm}0.003$ & $0.041{\pm}0.001$ & $0.044{\pm}0.002$ \\
\midrule
\multirow{3}{*}{ETTh2}
& 12.5\% & $0.034{\pm}0.001$ & $0.035{\pm}0.002$ & $0.041{\pm}0.003$ & $0.037{\pm}0.001$ & $0.201{\pm}0.002$ & $0.207{\pm}0.003$ & $0.030{\pm}0.000$ & $0.029{\pm}0.001$ \\
& 25\%   & $0.038{\pm}0.002$ & $0.035{\pm}0.003$ & $0.041{\pm}0.001$ & $0.039{\pm}0.002$ & $0.217{\pm}0.003$ & $0.214{\pm}0.001$ & $0.022{\pm}0.002$ & $0.027{\pm}0.003$ \\
& 50\%   & $0.051{\pm}0.003$ & $0.050{\pm}0.001$ & $0.007{\pm}0.000$ & $0.007{\pm}0.001$ & $0.371{\pm}0.002$ & $0.363{\pm}0.003$ & $0.015{\pm}0.001$ & $0.021{\pm}0.002$ \\
\midrule
\multirow{3}{*}{ETTm1}
& 12.5\% & $0.013{\pm}0.001$ & $0.011{\pm}0.000$ & $0.020{\pm}0.002$ & $0.022{\pm}0.003$ & $0.239{\pm}0.001$ & $0.225{\pm}0.002$ & $0.012{\pm}0.001$ & $0.017{\pm}0.003$ \\
& 25\%   & $0.016{\pm}0.001$ & $0.013{\pm}0.001$ & $0.009{\pm}0.000$ & $0.015{\pm}0.002$ & $0.260{\pm}0.003$ & $0.251{\pm}0.001$ & $0.017{\pm}0.003$ & $0.015{\pm}0.002$ \\
& 50\%   & $0.035{\pm}0.003$ & $0.035{\pm}0.001$ & $0.007{\pm}0.001$ & $0.008{\pm}0.000$ & $0.188{\pm}0.002$ & $0.182{\pm}0.003$ & $0.022{\pm}0.001$ & $0.026{\pm}0.002$ \\
\midrule
\multirow{3}{*}{ETTm2}
& 12.5\% & $0.011{\pm}0.000$ & $0.010{\pm}0.000$ & $0.012{\pm}0.002$ & $0.011{\pm}0.001$ & $0.182{\pm}0.003$ & $0.177{\pm}0.002$ & $0.004{\pm}0.000$ & $0.004{\pm}0.001$ \\
& 25\%   & $0.018{\pm}0.002$ & $0.017{\pm}0.001$ & $0.011{\pm}0.003$ & $0.014{\pm}0.002$ & $0.157{\pm}0.001$ & $0.155{\pm}0.003$ & $0.005{\pm}0.001$ & $0.004{\pm}0.000$ \\
& 50\%   & $0.020{\pm}0.003$ & $0.021{\pm}0.002$ & $0.008{\pm}0.000$ & $0.008{\pm}0.001$ & $0.161{\pm}0.002$ & $0.163{\pm}0.003$ & $0.004{\pm}0.001$ & $0.004{\pm}0.000$ \\
\midrule
\multirow{3}{*}{Weather}
& 12.5\% & $0.021{\pm}0.002$ & $0.018{\pm}0.001$ & $0.037{\pm}0.003$ & $0.037{\pm}0.000$ & $0.074{\pm}0.002$ & $0.072{\pm}0.001$ & $0.009{\pm}0.000$ & $0.008{\pm}0.001$ \\
& 25\%   & $0.023{\pm}0.003$ & $0.018{\pm}0.002$ & $0.015{\pm}0.001$ & $0.017{\pm}0.000$ & $0.077{\pm}0.003$ & $0.078{\pm}0.002$ & $0.010{\pm}0.001$ & $0.009{\pm}0.003$ \\
& 50\%   & $0.031{\pm}0.001$ & $0.026{\pm}0.003$ & $0.011{\pm}0.002$ & $0.010{\pm}0.001$ & $0.096{\pm}0.003$ & $0.094{\pm}0.002$ & $0.008{\pm}0.000$ & $0.008{\pm}0.001$ \\
\midrule
\multirow{3}{*}{ECL}
& 12.5\% & $0.074{\pm}0.001$ & $0.071{\pm}0.001$ & $0.009{\pm}0.000$ & $0.009{\pm}0.002$ & $0.101{\pm}0.003$ & $0.103{\pm}0.001$ & $0.058{\pm}0.002$ & $0.062{\pm}0.003$ \\
& 25\%   & $0.081{\pm}0.003$ & $0.080{\pm}0.002$ & $0.010{\pm}0.001$ & $0.009{\pm}0.000$ & $0.112{\pm}0.003$ & $0.105{\pm}0.002$ & $0.071{\pm}0.001$ & $0.072{\pm}0.003$ \\
& 50\%   & $0.095{\pm}0.001$ & $0.093{\pm}0.003$ & $0.008{\pm}0.000$ & $0.008{\pm}0.001$ & $0.148{\pm}0.002$ & $0.133{\pm}0.003$ & $0.093{\pm}0.002$ & $0.082{\pm}0.001$ \\
\bottomrule
\end{tabular}
}
\end{table*}

\begin{table*}[h]
\centering
\caption{Accuracy results (\%) on the classification task, comparing the original models (Ori) and TokenDecouple (TD) on pattern-centric tasks. The results are obtained from 5 random seeds.}
\label{tab:classification_acc}
\resizebox{\textwidth}{!}{
\begin{tabular}{l|cc|cc|cc|cc}
\toprule
\multirow{2}{*}{\textbf{Dataset}} 
& \multicolumn{2}{c|}{\textbf{OFA}} 
& \multicolumn{2}{c|}{\textbf{TimeLLM}} 
& \multicolumn{2}{c|}{\textbf{CALF}} 
& \multicolumn{2}{c}{\textbf{S\textsuperscript{2}IP}} \\
\cmidrule(lr){2-3} \cmidrule(lr){4-5} \cmidrule(lr){6-7} \cmidrule(lr){8-9}
& \textbf{Ori} & \textbf{TD} 
& \textbf{Ori} & \textbf{TD} 
& \textbf{Ori} & \textbf{TD} 
& \textbf{Ori} & \textbf{TD} \\
\midrule
JapaneseVowels 
& $98.2{\pm}0.07$ & $98.2{\pm}0.09$ 
& $30.1{\pm}0.06$ & $31.2{\pm}0.10$ 
& $95.9{\pm}0.08$ & $95.3{\pm}0.05$ 
& $93.6{\pm}0.04$ & $93.8{\pm}0.07$ \\
Handwriting 
& $35.2{\pm}0.09$ & $35.9{\pm}0.06$ 
& $12.5{\pm}0.08$ & $12.3{\pm}0.05$ 
& $27.1{\pm}0.10$ & $28.9{\pm}0.07$ 
& $27.7{\pm}0.06$ & $30.4{\pm}0.11$ \\
Heartbeat 
& $76.8{\pm}0.08$ & $77.1{\pm}0.05$ 
& $72.9{\pm}0.10$ & $72.4{\pm}0.07$ 
& $78.7{\pm}0.06$ & $77.4{\pm}0.09$ 
& $72.1{\pm}0.11$ & $72.7{\pm}0.08$ \\
EthanolConcentration 
& $34.8{\pm}0.06$ & $35.0{\pm}0.10$ 
& $31.2{\pm}0.09$ & $32.9{\pm}0.07$ 
& $29.6{\pm}0.05$ & $29.8{\pm}0.11$ 
& $29.7{\pm}0.08$ & $29.4{\pm}0.06$ \\
PEMS-SF 
& $84.2{\pm}0.10$ & $85.5{\pm}0.07$ 
& $25.1{\pm}0.06$ & $27.8{\pm}0.11$ 
& $78.9{\pm}0.09$ & $78.5{\pm}0.05$ 
& $82.8{\pm}0.08$ & $82.5{\pm}0.10$ \\
\bottomrule
\end{tabular}
}
\end{table*}

\begin{table*}[h]
\centering
\caption{F1-score results on the anomaly detection task, comparing the original models (Ori) and TokenDecouple (TD) on pattern-centric tasks. The results are obtained from 5 random seeds.}
\label{tab:anomaly_f1}
\resizebox{\textwidth}{!}{
\begin{tabular}{l|cc|cc|cc|cc}
\toprule
\multirow{2}{*}{\textbf{Dataset}} 
& \multicolumn{2}{c|}{\textbf{OFA}} 
& \multicolumn{2}{c|}{\textbf{TimeLLM}} 
& \multicolumn{2}{c|}{\textbf{CALF}} 
& \multicolumn{2}{c}{\textbf{S\textsuperscript{2}IP}} \\
\cmidrule(lr){2-3} \cmidrule(lr){4-5} \cmidrule(lr){6-7} \cmidrule(lr){8-9}
& \textbf{Ori} & \textbf{TD} 
& \textbf{Ori} & \textbf{TD} 
& \textbf{Ori} & \textbf{TD} 
& \textbf{Ori} & \textbf{TD} \\
\midrule
SMD  
& $85.74{\pm}0.09$ & $85.79{\pm}0.11$ 
& $84.12{\pm}0.08$ & $84.60{\pm}0.12$ 
& $86.54{\pm}0.10$ & $87.08{\pm}0.07$ 
& $83.34{\pm}0.13$ & $83.55{\pm}0.09$ \\
MSL  
& $85.11{\pm}0.10$ & $85.68{\pm}0.08$ 
& $69.76{\pm}0.08$ & $70.12{\pm}0.09$ 
& $84.68{\pm}0.11$ & $84.62{\pm}0.07$ 
& $80.22{\pm}0.12$ & $80.96{\pm}0.10$ \\
SMAP 
& $70.17{\pm}0.12$ & $72.32{\pm}0.09$ 
& $70.44{\pm}0.10$ & $71.18{\pm}0.08$ 
& $68.60{\pm}0.11$ & $68.87{\pm}0.11$ 
& $71.85{\pm}0.07$ & $72.26{\pm}0.10$ \\
SWaT 
& $93.09{\pm}0.08$ & $92.86{\pm}0.12$ 
& $91.63{\pm}0.11$ & $92.48{\pm}0.09$ 
& $94.56{\pm}0.07$ & $94.91{\pm}0.10$ 
& $83.46{\pm}0.13$ & $85.11{\pm}0.08$ \\
PSM  
& $97.76{\pm}0.07$ & $97.35{\pm}0.10$ 
& $92.46{\pm}0.12$ & $93.33{\pm}0.08$ 
& $96.24{\pm}0.09$ & $96.89{\pm}0.11$ 
& $95.61{\pm}0.12$ & $96.54{\pm}0.07$ \\
\bottomrule
\end{tabular}
}
\end{table*}

\begin{table*}[h]
\centering
\caption{MAE results on the forecasting task under two prediction horizons ($H=192,720$), comparing the original models (Ori) and TokenDecouple (TD) on pattern-centric tasks. The results are obtained from 5 random seeds.}
\label{tab:forecasting_mae}
\resizebox{\textwidth}{!}{
\begin{tabular}{ll|cc|cc|cc|cc}
\toprule
\multirow{2}{*}{\textbf{Dataset}} 
& \multirow{2}{*}{\textbf{Horizon}} 
& \multicolumn{2}{c|}{\textbf{OFA}} 
& \multicolumn{2}{c|}{\textbf{TimeLLM}} 
& \multicolumn{2}{c|}{\textbf{CALF}} 
& \multicolumn{2}{c}{\textbf{S\textsuperscript{2}IP}} \\
\cmidrule(lr){3-4} \cmidrule(lr){5-6} \cmidrule(lr){7-8} \cmidrule(lr){9-10}
& & \textbf{Ori} & \textbf{TD} 
  & \textbf{Ori} & \textbf{TD} 
  & \textbf{Ori} & \textbf{TD} 
  & \textbf{Ori} & \textbf{TD} \\
\midrule
\multirow{2}{*}{Traffic}
& 192 & $0.397{\pm}0.001$ & $0.386{\pm}0.005$ & $0.374{\pm}0.002$ & $0.368{\pm}0.006$ & $0.437{\pm}0.004$ & $0.432{\pm}0.000$ & $0.391{\pm}0.003$ & $0.395{\pm}0.001$ \\
& 720 & $0.435{\pm}0.006$ & $0.422{\pm}0.002$ & $0.430{\pm}0.004$ & $0.425{\pm}0.001$ & $0.485{\pm}0.003$ & $0.484{\pm}0.006$ & $0.416{\pm}0.002$ & $0.414{\pm}0.005$ \\
\midrule
\multirow{2}{*}{Solar}
& 192 & $0.193{\pm}0.004$ & $0.195{\pm}0.000$ & $0.160{\pm}0.006$ & $0.152{\pm}0.002$ & $0.248{\pm}0.001$ & $0.251{\pm}0.005$ & $0.179{\pm}0.003$ & $0.175{\pm}0.006$ \\
& 720 & $0.202{\pm}0.001$ & $0.212{\pm}0.004$ & $0.190{\pm}0.005$ & $0.179{\pm}0.003$ & $0.251{\pm}0.006$ & $0.244{\pm}0.002$ & $0.203{\pm}0.000$ & $0.196{\pm}0.004$ \\
\midrule
\multirow{2}{*}{PEMS04}
& 192 & $0.458{\pm}0.002$ & $0.447{\pm}0.006$ & $0.453{\pm}0.001$ & $0.455{\pm}0.004$ & $0.783{\pm}0.005$ & $0.775{\pm}0.003$ & $0.564{\pm}0.006$ & $0.561{\pm}0.001$ \\
& 720 & $0.568{\pm}0.004$ & $0.570{\pm}0.001$ & $0.518{\pm}0.006$ & $0.514{\pm}0.002$ & $0.775{\pm}0.000$ & $0.764{\pm}0.005$ & $0.597{\pm}0.003$ & $0.593{\pm}0.006$ \\
\midrule
\multirow{2}{*}{PEMS08}
& 192 & $0.515{\pm}0.005$ & $0.502{\pm}0.003$ & $0.525{\pm}0.000$ & $0.507{\pm}0.006$ & $0.844{\pm}0.002$ & $0.842{\pm}0.005$ & $0.703{\pm}0.001$ & $0.692{\pm}0.004$ \\
& 720 & $0.617{\pm}0.002$ & $0.594{\pm}0.006$ & $0.589{\pm}0.003$ & $0.568{\pm}0.001$ & $0.873{\pm}0.004$ & $0.868{\pm}0.006$ & $0.764{\pm}0.005$ & $0.758{\pm}0.002$ \\
\midrule
\multirow{2}{*}{Exchange}
& 192 & $0.181{\pm}0.006$ & $0.165{\pm}0.001$ & $0.208{\pm}0.003$ & $0.194{\pm}0.005$ & $0.163{\pm}0.002$ & $0.152{\pm}0.004$ & $0.205{\pm}0.001$ & $0.201{\pm}0.006$ \\
& 720 & $0.841{\pm}0.004$ & $0.814{\pm}0.002$ & $0.897{\pm}0.006$ & $0.882{\pm}0.001$ & $0.842{\pm}0.003$ & $0.820{\pm}0.005$ & $0.993{\pm}0.002$ & $0.978{\pm}0.004$ \\
\bottomrule
\end{tabular}
}
\end{table*}

\begin{table*}[h]
\centering
\caption{MSE results on the forecasting task under two prediction horizons ($H=192,720$), comparing the original models (Ori) and TokenDecouple (TD) on pattern-centric tasks. The results are obtained from 5 random seeds.}
\label{tab:forecasting_mse}
\resizebox{\textwidth}{!}{
\begin{tabular}{ll|cc|cc|cc|cc}
\toprule
\multirow{2}{*}{\textbf{Dataset}} 
& \multirow{2}{*}{\textbf{Horizon}} 
& \multicolumn{2}{c|}{\textbf{OFA}} 
& \multicolumn{2}{c|}{\textbf{TimeLLM}} 
& \multicolumn{2}{c|}{\textbf{CALF}} 
& \multicolumn{2}{c}{\textbf{S\textsuperscript{2}IP}} \\
\cmidrule(lr){3-4} \cmidrule(lr){5-6} \cmidrule(lr){7-8} \cmidrule(lr){9-10}
& & \textbf{Ori} & \textbf{TD} 
  & \textbf{Ori} & \textbf{TD} 
  & \textbf{Ori} & \textbf{TD} 
  & \textbf{Ori} & \textbf{TD} \\
\midrule
\multirow{2}{*}{Traffic}
& 192 & $0.283{\pm}0.001$ & $0.277{\pm}0.002$ & $0.265{\pm}0.006$ & $0.267{\pm}0.003$ & $0.265{\pm}0.005$ & $0.261{\pm}0.001$ & $0.280{\pm}0.007$ & $0.284{\pm}0.004$ \\
& 720 & $0.297{\pm}0.005$ & $0.285{\pm}0.003$ & $0.298{\pm}0.002$ & $0.293{\pm}0.006$ & $0.294{\pm}0.004$ & $0.295{\pm}0.001$ & $0.288{\pm}0.008$ & $0.290{\pm}0.002$ \\
\midrule
\multirow{2}{*}{Solar}
& 192 & $0.266{\pm}0.003$ & $0.263{\pm}0.001$ & $0.237{\pm}0.004$ & $0.233{\pm}0.006$ & $0.256{\pm}0.002$ & $0.258{\pm}0.005$ & $0.251{\pm}0.001$ & $0.247{\pm}0.003$ \\
& 720 & $0.279{\pm}0.004$ & $0.281{\pm}0.006$ & $0.256{\pm}0.002$ & $0.251{\pm}0.003$ & $0.270{\pm}0.005$ & $0.263{\pm}0.004$ & $0.277{\pm}0.006$ & $0.273{\pm}0.002$ \\
\midrule
\multirow{2}{*}{PEMS04}
& 192 & $0.362{\pm}0.002$ & $0.358{\pm}0.003$ & $0.361{\pm}0.004$ & $0.364{\pm}0.001$ & $0.517{\pm}0.006$ & $0.509{\pm}0.004$ & $0.442{\pm}0.002$ & $0.442{\pm}0.003$ \\
& 720 & $0.414{\pm}0.006$ & $0.413{\pm}0.004$ & $0.398{\pm}0.003$ & $0.402{\pm}0.002$ & $0.504{\pm}0.005$ & $0.501{\pm}0.006$ & $0.476{\pm}0.004$ & $0.477{\pm}0.001$ \\
\midrule
\multirow{2}{*}{PEMS08}
& 192 & $0.373{\pm}0.004$ & $0.370{\pm}0.006$ & $0.377{\pm}0.003$ & $0.369{\pm}0.005$ & $0.508{\pm}0.006$ & $0.511{\pm}0.002$ & $0.422{\pm}0.004$ & $0.418{\pm}0.003$ \\
& 720 & $0.412{\pm}0.006$ & $0.407{\pm}0.005$ & $0.402{\pm}0.004$ & $0.395{\pm}0.003$ & $0.510{\pm}0.006$ & $0.510{\pm}0.004$ & $0.469{\pm}0.005$ & $0.463{\pm}0.002$ \\
\midrule
\multirow{2}{*}{Exchange}
& 192 & $0.295{\pm}0.003$ & $0.290{\pm}0.002$ & $0.331{\pm}0.004$ & $0.322{\pm}0.001$ & $0.281{\pm}0.005$ & $0.277{\pm}0.003$ & $0.324{\pm}0.004$ & $0.321{\pm}0.006$ \\
& 720 & $0.688{\pm}0.002$ & $0.667{\pm}0.004$ & $0.735{\pm}0.005$ & $0.718{\pm}0.003$ & $0.689{\pm}0.006$ & $0.679{\pm}0.004$ & $0.792{\pm}0.005$ & $0.782{\pm}0.003$ \\
\bottomrule
\end{tabular}
}
\end{table*}

\clearpage
\onecolumn
\section{Algorithmic of TokenDecouple}

\begin{table*}[h]
\caption{Algorithmic of {TokenDecouple}. The method decouples TS tokens and prompt tokens via frequency-aware TS token merging and pyramidal prompt token compression.}
\centering
\small
\setlength{\tabcolsep}{6pt}
\renewcommand{\arraystretch}{1.2}
\begin{tabular}{p{0.22\linewidth} p{0.74\linewidth}}
\toprule
\textbf{Module} & \textbf{Description} \\
\midrule

\textbf{TS Token Merging} &
\textbf{Input:} TS tokens $\mathbf{X}=\{\mathbf{x}_i\}_{i=1}^{N}$, $\mathbf{x}_i\in\mathbb{R}^D$. \\
&
1) Compute DFT along the token axis: $\mathbf{F}=\mathrm{DFT}(\mathbf{X})$, and keep non-negative frequencies $k=0,\ldots,K-1$, $K=\lfloor N/2\rfloor+1$. \\
&
2) For each sample $b$ and frequency $k$, compute magnitudes $\hat{F}_{b,i,k}=|F_{b,i,k}|$, and select the minimal active set $\mathcal{I}_{b,k}$ whose cumulative magnitude exceeds $95\%$ of the total; set $n_{b,k}=|\mathcal{I}_{b,k}|$. \\
&
3) For each $i\in\mathcal{I}_{b,k}$, estimate spectral identifiability via Fisher information
$s_{b,i,k}=\mathrm{tr}(-\mathbb{E}[\nabla^2_{\boldsymbol{\theta}_{b,k}}\ell_{b,i}])$, and aggregate
$S_{b,k}=\frac{1}{n_{b,k}}\sum_{i\in\mathcal{I}_{b,k}} s_{b,i,k}$. \\
&
4) Convert $S_{b,k}$ to merging strength $r_{b,k}$ via a sigmoid, and determine the adaptive merging budget
$m_{b,k}=\lceil(1-r_{b,k})\,n_{b,k}\rceil$. \\
&
5) Build a sparse local affinity graph for $|i-j|<r_{\text{loc}}$ with
$A_{i,j}=\exp(\mathrm{sim}(\phi(|F_{i,k}|),\phi(|F_{j,k}|))/\tau)$, and partition $\mathcal{I}_{b,k}$ into $m_{b,k}$ local groups $\{\mathcal{G}^{(u)}_{b,k}\}$. \\
&
6) For each group $\mathcal{G}^{(u)}_{b,k}$, compute weighted merging
$\tilde{\mathbf{F}}_{u,k}=\sum_{i\in\mathcal{G}^{(u)}_{b,k}}\omega_i^{(u,k)}\mathbf{F}_{i,k}$,
where $\omega_i^{(u,k)}$ is normalized by intra-group affinities. \\
&
7) Apply IDFT over $\{\tilde{\mathbf{F}}_{u,k}\}_{k=0}^{K-1}$ to obtain merged TS tokens $\tilde{\mathbf{X}}$, and add penalty
$L_{\mathrm{pen}}=\lambda\frac{\sum_k m_{b,k}}{\sum_k n_{b,k}}$. \\
\midrule

\textbf{Prompt Tokens Compression} &
\textbf{Input:} prompt tokens $\mathbf{H}_P^{(l)}\in\mathbb{R}^{P_l\times D}$ at layer $l$. \\
\textbf{} &
1) Reduce prompt length with pyramidal decay: $P_{l+1}=\lceil 0.25\,P_l\rceil$. \\
&
2) Learn a row-normalized compression matrix $\mathbf{W}^{(l)}\in\mathbb{R}^{P_{l+1}\times P_l}$ and compute compressed prompts
$\mathbf{S}^{(l)}=\mathbf{W}^{(l)}\mathbf{H}_P^{(l)}$. \\
&
3) Compute a global residual summary $\mathbf{r}^{(l)}=\mathrm{MeanPool}(\mathbf{H}_P^{(l)})$. \\
&
4) Update prompt tokens for the next layer:
$\mathbf{H}_P^{(l+1)}=\mathbf{S}^{(l)}+\mathbf{1}_{P_{l+1}}\mathbf{r}^{(l)}$. \\
&
5) Repeat until layer $L$, yielding rapidly decaying prompt counts (25\%, 6.3\%, 1.6\%, 0.4\%, $\ldots$) and negligible overhead in deep layers. \\
\bottomrule
\end{tabular}

\label{tab:tokendecouple}
\end{table*}

\twocolumn

\section{Prompt Templates}
To ensure a consistent input format across different time-series tasks, we provide the prompt templates used in our experiments. 
Fig.~\ref{fig:prompt_template_imputation}--\ref{fig:prompt_template_reasoning} illustrate the templates for five  tasks: forecasting, imputation, classification, anomaly detection, and reasoning. 
For pattern-centric tasks, including forecasting, imputation, classification, and anomaly detection, the prompts mainly specify the input sequence, task objective, and expected output format. 
For the reasoning task, the prompt further incorporates contextual information and candidate answers, requiring the model to jointly interpret temporal evidence and domain-specific semantics. 
These templates standardize the textual interface while allowing each task to preserve its own output structure.
\begin{figure}[h]
\begin{tcolorbox}[
    title=\textbf{Example of the Imputation task},
    colback=gray!5,
    colframe=gray!60,
    boxrule=0.6pt,
    arc=2pt,
    left=5pt,
    right=5pt,
    top=5pt,
    bottom=5pt
]
\textbf{Instruction:} You are an expert in time series imputation. 
Given an incomplete time series, please infer the missing values based on the observed temporal evidence.

\textbf{Input Time Series:} 
[0.541,\ 0.682,\ 1.322,\ \texttt{[MASK]}, ..., \ 1.087,\ 0.936,\ \texttt{[MASK]},\ 0.952]

\textbf{Task Description:} 
Given a time series with missing observations, recover the missing values by using the surrounding temporal context, local continuity, and global temporal patterns.

\textbf{Missing Pattern:} 
The missing values are marked as \texttt{[MASK]} in the input sequence.

\textbf{Output:} 
Please output the imputed values for all masked positions in their original order.
\end{tcolorbox}
\caption{Prompt template for the Imputation task.}
\label{fig:prompt_template_imputation}
\end{figure}

\begin{figure}[h]
\begin{tcolorbox}[
    title=\textbf{Example of the Classification task},
    colback=gray!5,
    colframe=gray!60,
    boxrule=0.6pt,
    arc=2pt,
    left=5pt,
    right=5pt,
    top=5pt,
    bottom=5pt
]
\textbf{Instruction:} You are an expert in time series classification. 
Given an input time series, please assign it to the correct predefined class according to its temporal dynamics and discriminative patterns.

\textbf{Input Time Series:} 
[
[0.312,\ -0.184,\ 0.527,\ ...,\ 0.091], 
[0.286,\ -0.207,\ 0.498,\ ...,\ 0.116], 
[0.241,\ -0.263,\ 0.462,\ ...,\ 0.154], 
..., 
[0.105,\ -0.391,\ 0.336,\ ...,\ 0.218]
]

\textbf{Task Description:} 
Given a multivariate time series instance, identify its class label from the predefined label set by analyzing temporal evolution, variable interactions, and class-specific dynamic patterns.

\textbf{Candidate Classes:}

(A) Class 1 \\
(B) Class 2 \\
(C) Class 3 \\
(D) Class 4 \\
(E) Class 5

\textbf{Output:} 
Please output the correct class label.
\end{tcolorbox}
\caption{Prompt template for the Classification task.}
\label{fig:prompt_template_classification}
\end{figure}

\begin{figure}[h]
\begin{tcolorbox}[
    title=\textbf{Example of the Anomaly Detection task},
    colback=gray!5,
    colframe=gray!60,
    boxrule=0.6pt,
    arc=2pt,
    left=5pt,
    right=5pt,
    top=5pt,
    bottom=5pt
]
\textbf{Instruction:} You are an expert in time series anomaly detection. 
Given an input time series, please identify whether each time step is normal or anomalous.

\textbf{Input Time Series:} 
[0.318,\ 0.336,\ 0.351,\ 0.329,\ ..., \ 0.374,\ 2.186,\ 0.392,\ 0.365]

\textbf{Task Description:} 
Given the input sequence, perform point-wise anomaly detection by assigning an anomaly label to each time step. 
A label of 0 indicates a normal point, while a label of 1 indicates an anomalous point. 
Anomalies may correspond to sudden spikes, abrupt drops, abnormal fluctuations, distribution shifts, or deviations from normal temporal patterns.

\textbf{Label Definition:}

0: Normal time point \\
1: Anomalous time point

\textbf{Output:} 
Please output a binary label sequence with the same length as the input time series, where each position corresponds to the anomaly label of the corresponding time step.
\end{tcolorbox}
\caption{Prompt template for the Anomaly Detection task.}
\label{fig:prompt_template_anomaly}
\end{figure}

\begin{figure}[h]
\begin{tcolorbox}[
    title=\textbf{Example of the Forecasting task},
    colback=gray!5,
    colframe=gray!60,
    boxrule=0.6pt,
    arc=2pt,
    left=5pt,
    right=5pt,
    top=5pt,
    bottom=5pt
]
\textbf{Instruction:} You are an expert in time series forecasting. 
Given a historical time series, please predict its future values based on the observed temporal evidence.

\textbf{Input Time Series:} 
Past 512 time steps: 
[0.214,\ 0.287,\ 0.356,\ 0.421,\ ..., \ 0.764,\ 0.812,\ 0.857,\ 0.903]

\textbf{Task Description:} 
Given the observed historical sequence of length 512, forecast the future sequence of length 192 by considering temporal continuity, trend, periodicity, and recent fluctuations.

\textbf{Prediction Horizon:} 
Please predict the next 192 time steps.

\textbf{Output:} 
Please output the predicted future time series values in chronological order:
$[\hat{x}_{513},\ \hat{x}_{514},\ \ldots,\ \hat{x}_{704}]$.
\end{tcolorbox}
\caption{Prompt template for the Forecasting task.}
\label{fig:prompt_template_forecasting}
\end{figure}

\begin{figure}[h]
\begin{tcolorbox}[
    title=\textbf{Example of the Reasoning task},
    colback=gray!5,
    colframe=gray!60,
    boxrule=0.6pt,
    arc=2pt,
    left=5pt,
    right=5pt,
    top=5pt,
    bottom=5pt
]
\textbf{Instruction:} You are an expert in time series analysis and temporal reasoning. 
Given a time series and its contextual information, please answer the multiple-choice question based on the temporal evidence, domain context, and candidate answers. Only one answer is correct.

\textbf{Input Time Series:} 
[0.124,\ 0.139,\ 0.151,\ 0.148,\ ..., \ 0.632,\ 1.847,\ 0.914,\ 0.386]

\textbf{Context:} 
The sequence records a physical signal collected from a transient electromagnetic event. 
A sharp radiation spike followed by subsequent signal variations may indicate a positive cloud-to-ground discharge event.

\textbf{Task Description:} 
Given the temporal pattern and contextual information, infer the most plausible event type. 
The answer should be derived from both the observed time series dynamics and the provided domain knowledge, rather than from the numerical pattern alone.

\textbf{Question:} 
Which event type is most consistent with the observed temporal signal and the provided context?

\textbf{Candidate Answers:}

(A) CG Positive discharge \\
(B) Background noise without transient event \\
(C) Periodic sensor drift \\
(D) Long-duration stable radiation source

\textbf{Output:} 
Please output the correct candidate answer.
\end{tcolorbox}
\caption{Prompt template for the Reasoning task.}
\label{fig:prompt_template_reasoning}
\end{figure}

\clearpage
\section{Hyperparameter Sensitivity Analysis}
\subsection{Effect of Regularization Weight}
\begin{figure}[h]
    \centering
    \includegraphics[width=\linewidth]{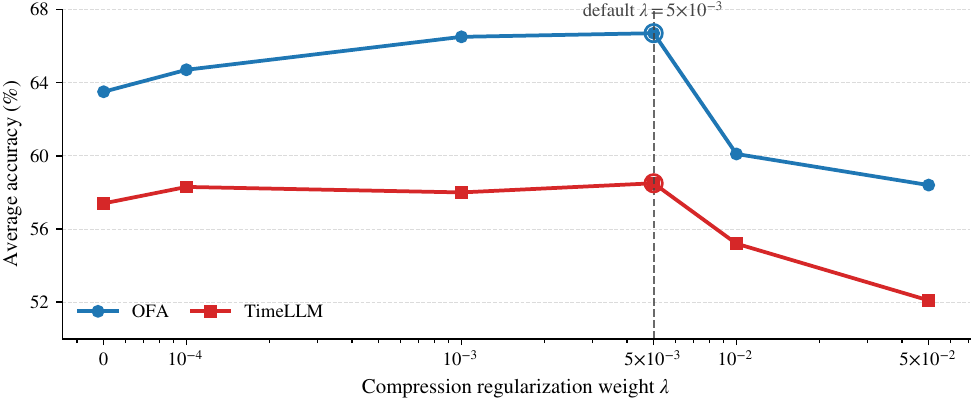}
    \caption{
    Sensitivity analysis of the compression regularization weight $\lambda$ on reasoning tasks. 
    }
    \label{fig:lambda_sensitivity}
\end{figure}

To examine the effect of the compression regularization weight $\lambda$, we conduct a sensitivity analysis on the reasoning tasks and report the average performance over six tasks. 
As shown in Fig.~\ref{fig:lambda_sensitivity}, $\lambda$ controls the trade-off between compactness and task performance. 
When $\lambda$ is too small, the compression constraint is weak and the model cannot fully exploit token redundancy. 
As $\lambda$ increases to $5\times10^{-3}$, both OFA and TimeLLM achieve their best or near-best performance, indicating that a moderate compactness constraint can remove redundant TS tokens while preserving useful frequency information. 
However, when $\lambda$ further increases to $10^{-2}$ or $5\times10^{-2}$, the performance drops noticeably, suggesting that overly strong compression may discard task-relevant temporal information. 
Therefore, we set $\lambda=5\times10^{-3}$ as the default value, which provides a favorable balance between efficiency and reasoning performance.

\subsection{Effect of Local Affinity Radius}
We further analyze the effect of the local affinity radius $r_{\text{loc}}$, which controls the neighborhood size used to construct the sparse local graph for TS token merging. 
As shown in Table~\ref{tab:rloc_sensitivity}, a small $r_{\text{loc}}$ restricts the searchable merging candidates, making the compression overly conservative and limiting the benefit of redundancy removal. 
Increasing $r_{\text{loc}}$ to a moderate value improves the average reasoning performance, indicating that a proper local neighborhood helps identify spectrally similar tokens and produces more reliable merged representatives. 
However, when $r_{\text{loc}}$ becomes too large, the performance starts to decline. 
This is because a large neighborhood may connect tokens with different spectral structures, increasing the group-wise spectral diameter and introducing larger merging errors. 
Therefore, we set $r_{\text{loc}}=4$ by default, which provides a stable balance between local spectral consistency and compression effectiveness.
\begin{table}[h]
\centering
\caption{
Sensitivity analysis of the local affinity radius $r_{\text{loc}}$ on reasoning tasks. 
}
% \small
\begin{tabular}{c|cc}
\toprule
$r_{\text{loc}}$ & OFA & TimeLLM \\
\midrule
1  & 64.1 & 57.6 \\
2  & 65.8 & 58.1 \\
4  & \textbf{66.7} & \textbf{58.5} \\
8  & 65.9 & 57.9 \\
16 & 62.8 & 55.6 \\
\bottomrule
\end{tabular}

\label{tab:rloc_sensitivity}
\end{table}
% \clearpage
% \input{sections/xiuding}
\end{document}